\documentclass{article}

\usepackage{longtable}
\usepackage{IEEEtrantools}
\usepackage{arxiv}
\usepackage{amsmath,amssymb}
\usepackage{float}
\usepackage[utf8]{inputenc} 
\usepackage[T1]{fontenc}    
\usepackage{hyperref}       
\usepackage{url}            
\usepackage{booktabs}       
\usepackage{amsfonts}       
\usepackage{nicefrac}       
\usepackage{microtype}      
\usepackage{lipsum}
\usepackage{graphicx}
 \usepackage{natbib}
\graphicspath{ {./images/} }
\newcommand{\E}{\mathbb{E}}

\newtheorem{theorem}{Theorem}
\newtheorem{proof}{proof}

\newtheorem{proposition}{Proposition}
\newtheorem{lemma}{Lemma}
\newtheorem{corollary}{Corollary}
\newtheorem{definition}{Definition}

\title{Theory Foundation of Physics-Enhanced Residual Learning}

\author{
 Shixiao Liang \\
  Department of Civil and Environmental Engineering\\
  University of Wisconsin-Madison\\
  Madison, WI, 53706 \\
  \texttt{sliang85@wisc.edu} \\
   \And
 Wang Chen \\
  Department of Civil Engineering\\
  The University of Hong Kong\\
  Hong Kong, China \\
  \texttt{wchen22@connect.hku.hk} \\
  \And
 Keke Long* \\
  Department of Civil and Environmental Engineering\\
  University of Wisconsin-Madison\\
  Madison, WI, 53706 \\
  \texttt{klong23@wisc.edu} \\
    \And
 Peng Zhang\\
  Department of Civil and Environmental Engineering\\
  University of Wisconsin-Madison\\
  Madison, WI, 53706 \\
  \texttt{pzhang257@wisc.edu} \\
      \And
 Xiaopeng Li*\\
  Department of Civil and Environmental Engineering\\
  University of Wisconsin-Madison\\
  Madison, WI, 53706 \\
  \texttt{xli2485@wisc.edu} \\
        \And
 Jintao Ke\\
  Department of Civil Engineering\\
  The University of Hong Kong\\
  Hong Kong, China \\
  \texttt{kejintao@hku.hk} \\
}

\begin{document}
\maketitle
\begin{abstract}
\textbf{Problem definition:} Intensive studies have been conducted in recent years to integrate neural networks with physics models to balance model accuracy and interpretability. One recently proposed approach, named \textbf{Physics-Enhanced Residual Learning} (PERL), is to use learning to estimate the residual between the physics model prediction and the ground truth. Numeral examples suggested that integrating such residual with physics models in PERL has three advantages: (1) a reduction in the number of required neural network parameters; (2) faster convergence rates; and (3) fewer training samples needed for the same computational precision. However, these numerical results lack theoretical justification and cannot be adequately explained. 

\textbf{Methodology:} This paper aims to explain these advantages of PERL from a theoretical perspective. We investigate a general class of problems with Lipschitz continuity properties. By examining the relationships between the bounds to the loss function and residual learning structure, this study rigorously proves a set of theorems explaining the three advantages of PERL. 

\textbf{Implications:} Several numerical examples in the context of automated vehicle trajectory prediction are conducted to illustrate the proposed theorems. The results confirm that, even with significantly fewer training samples, PERL consistently achieves higher accuracy than a pure neural network. These results demonstrate the practical value of PERL in real world autonomous driving applications where corner case data are costly or hard to obtain. PERL therefore improves predictive performance while reducing the amount of data required.

\end{abstract}


\section{Introduction}

In recent years, Neural Network (NN) models have received significant attention due to their remarkable predictive capabilities in transportation applications, including vehicle behavior prediction \citep{zhou2017recurrent,shi2022integrated}, traffic flow prediction \citep{do2019effective, kang2017short}, and behavioral choice modeling \citep{wang2020deep,wang2020deep1}. Although NNs excel at capturing complex nonlinear patterns inherent in real-world problems, their performance heavily depends on large and high-quality training datasets. As a result, when datasets suffer from low quality, missing data, or excessive noise, the training process can become less accurate, resulting in suboptimal model performance \citep{liu2024navigating,emmanuel2021survey}. Additionally, NNs often involve numerous parameters to effectively learn the high-dimensional data, leading to increased memory consumption and computational costs. Furthermore, the inherent black-box nature of NN models reduces interpretability, limiting their reliability in safety-critical applications. 


To overcome these limitations, researchers have proposed integrating physics models with NNs, which has shown great potential in various fields such as applied mathematics \citep{wang2024pinn, hu2024tackling}, material science \citep{chew2024advancing, faroughi2024physics}, and transportation science \citep{shi2023physics, pan2024ro, mo2021physics}. Prior approaches incorporate physical knowledge as regularization during model training. For instance, Physics-Informed Neural Networks (PINNs) penalize the mismatch between NN predictions and physical equations by adding this mismatch into the training loss~\citep{raissi2017physics, long2024PINN}, and Physics-Regularized Gaussian Processes (PRGP)  penalize deviations from physical laws across the input domain~\citep{yuan2021macroscopic}. Although these approaches effectively integrate physical knowledge into NNs, they primarily focus on imposing physical constraints during training. In contrast, the idea of residual learning offers an alternative perspective. 
Residual learning, which has been widely adopted in modern deep learning architectures such as ResNet~\citep{he2016deep}, recurrent NNs~\citep{hochreiter1997long}, and Transformer models~\citep{vaswani2017attention}, emphasizes preserving the main trend of a target function while concentrating learning efforts on the smaller residual. This strategy has been shown to facilitate optimization, improve convergence, and enhance generalization across various tasks \citep{dona2022constrainedICLR}, including numerical methods \citep{welch1995introduction,trottenberg2001multigrid},robot control \citep{he2025asap}, time series forecasting in supply-demand network \citep{said2019deep} and crowd flow \citep{zhang2017deep}.

\begin{figure}[H]
    \centering
    \includegraphics[width=0.95\textwidth]{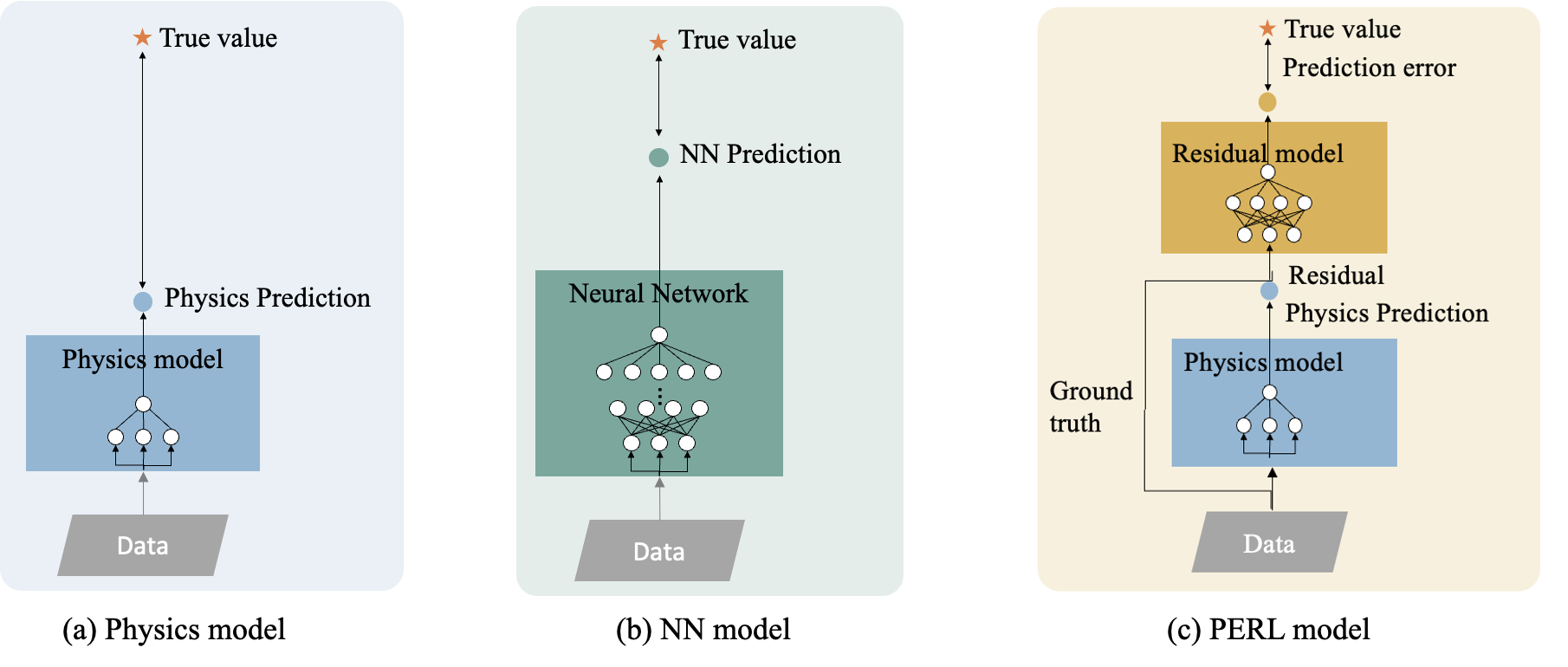}
    \caption{General structure of PERL compared with physics model and NN model \citep{long2023physics}}
    \label{fig:General_structure_of_PERL.png}
\end{figure}

Motivated by these advantages, \citet{long2023physics} proposed the Physics-Enhanced Residual Learning (PERL) framework, as illustrated in Figure~\ref{fig:General_structure_of_PERL.png}c. Compared to the traditional physics model (Figure~\ref{fig:General_structure_of_PERL.png}a) and the pure NN model (Figure~\ref{fig:General_structure_of_PERL.png}b), PERL integrates the strengths of both approaches. In PERL, a physics model is first employed to generate initial predictions that capture the dominant trend of the system, providing reasonable but potentially less accurate results. A neural network is then trained to predict the residual, which is the discrepancy between the initial physics prediction and the ground truth data, serving as a correction term. By concentrating the learning task on residual components, PERL simplifies NN training, improves prediction efficiency and accuracy, and preserves the interpretability afforded by the physics model.

 Before rigorous analysis, we can intuitively have the following observations based on the PERL framework as shown above. Generally, the physical model can achieve relatively accurate predictions with only several parameters. Thus, after processed by the physical model, the architecture of the residual learning model can be significantly lighter than a pure NN, as it only need to predict the residuals. As a result, the number of parameters of PERL can be reduced. Besides, the physical model is concluded based on physical principles or calibrated experiment results, which can reflect the data trend. Hence, the fluctuations of the residuals can be much smoother than the original data, making it easier for the residual learning model to find the optimal point \citep{raissi2019physics}. This leads to a faster convergence rate compared with learning the data structure from the beginning. In addition, when we cannot access large-scale data, the PREL can ensure a higher accuracy compared with a pure NN, since the physical model can output relatively accurate prediction results. In a nutshell, compared to traditional NNs, the PERL architecture effectively addresses the following challenges:

1. High Model Complexity:
Traditional models often require a large number of parameters, increasing computational costs and overfitting risks.

2. Prolonged Training Time: Large fluctuations in the target function can result in gradient variations, extending the time required for convergence.

3. Inadequate Training Due to Insufficient Data: Many scenarios suffer from a lack of sufficient data, resulting in poorly trained models with limited generalization ability.

Despite its practical success \citep{zhang2024online}, its theoretical foundations remain underdeveloped. In contrast, frameworks like PINNs and PRML are supported by systematic theoretical analyses, including convergence guarantees and error estimates~\citep{mishra2022estimates, yuan2021macroscopic}. This comparison highlights the critical need for a rigorous theoretical foundation for PERL as well. Establishing such a foundation is essential for validating the advantages of PERL: reduced model complexity, accelerated training convergence, and improved generalization under limited data. 
Because theoretical guarantees are crucial for enhancing the interpretability and reliability of machine learning models \citep{lipton2018mythos, doshi2017towards}. Therefore, a more comprehensive theoretical exploration of PERL is necessary. To fill these gaps, we aim to provide a systematic theoretical analysis to substantiate these benefits from the perspectives of NN parameter size, convergence rate, and statistical error bound.

\subsection{Parameter size}

NNs are parameterized models whose complexity and computational cost are determined by the number of trainable parameters, including weights and biases in each neurons \citep{gurney2018introduction, rumelhart1986learning, moody1991effective,han2015learning}. The total number of parameters directly influences both predictive capacity and resource requirements. While previous studies have empirically demonstrated that the PERL model can achieve comparable performance with significantly fewer parameters, the theoretical explanation for this phenomenon has not yet been established. This lack of explanation impacts the interpretability of PERL and makes it difficult to accurately estimate the number of parameters when using the PERL approach.

\subsection{Convergence rate}

The stochastic gradient descent (SGD) method is fundamental for optimizing NNs, updating weights and biases along the negative gradient of the loss function to minimize prediction errors \citep{ruder2016overview}. The effectiveness of gradient descent in NNs depends not only on the choice of the optimization algorithm but also on the network architecture itself \citep{shalev2014understanding}. Different architectures can lead to different convergence rates. Thus, understanding convergence rate behavior is crucial for evaluating how efficiently gradient descent works in NNs and how PERL achieves faster convergence compared to traditional architectures. Convergence rate bounds provide a theoretical framework for analyzing optimization speed, offering insights into why PERL model accelerates convergence speed compared to traditional NNs.

\subsection{Error bound}
In statistics, generalization error and estimation error are commonly used to assess the performance of NNs \citep{yarotsky2017error, shultzman2023generalization}. Estimation error arises from using limited training data to estimate model parameters, reflecting the difference between estimated parameters and their true optimal values. It is closely related to the model's complexity and the amount of available data. Generalization error, on the other hand, refers to the expected error of the model on unseen data, measuring the average difference between the model's predictions and true values on new samples. Effectively bounding generalization and estimation errors provide a theoretical measure of model quality. From a statistical perspective, error bounds provide a theoretical framework for understanding how PERL achieves comparable performance with fewer training samples than traditional NNs. This indicates that PERL enhances learning efficiency, enabling effective model training with limited data.
\vspace{\baselineskip}

In summary, this paper's contribution lies in theoretically proving three advantages of PERL over traditional NNs from the three aforementioned perspectives. Each proof is supplemented with an example to illustrate the practical feasibility of our theories. Furthermore, to validate the theoretical results, we utilize the Ultra-AV dataset, a unified longitudinal trajectory dataset for automated vehicle,  \citep{zhou2024ultra} to compare the performance of PERL and traditional NNs. The experimental results strongly agree with the theorems, verifying the effectiveness and correctness of the theoretical findings.

The structure of this paper is as follows: In Section \ref{Review}, we introduce the PERL mathematical framework and two key assumptions. In Section \ref{Theory}, we show the proof of  three theoretical results, each illustrated by a simple example. In Section \ref{experiments}, we conduct three trajectory prediction experiments using real-world AV datasets to demonstrate the validity of the theories presented in Section \ref{Theory}. Section \ref{conclusion} concludes the paper and discusses potential future research directions.

\section{Methodology Review}
\label{Review}

\begin{center}
\begin{longtable}{@{}p{3.5cm} p{10cm}@{}}
\caption{Notation List}
\label{tab:Notations}\\
\hline
\multicolumn{2}{l}{\textbf{Parameters}}\\
\hline
\endfirsthead

\multicolumn{2}{c}\textsl{\small continued from previous page}\\
\hline
\multicolumn{2}{l}{\textbf{Parameters} (cont.)}\\
\hline
\endhead

\hline \multicolumn{2}{r}\textsl{\small continued from previous page}\\
\endfoot

\hline
\endlastfoot

\(\Omega\subset\mathbb{R}^d\)  
  & Compact domain of all possible states \(s\). \\[6pt]

\(s\)  
  & State vector in \(\Omega\subset\mathbb{R}^d\); contains system variables (e.g.\ speed, acceleration). \\[6pt]

\(S_{\text{Phy}}(s)\)  
  & Projection of \(s\) onto a lower‐dimensional physics subspace \(\mathcal{S}_{\text{Phy}}\). \\[6pt]

\(\theta^{\text{Phy}}\)  
  & Parameters of the physics model \(f^{\text{Phy}}\). \\[6pt]

\(\theta^{\text{RL}}\)  
  & Parameters of the residual‐learning model \(f^{\text{RL}}\). \\[6pt]

\(\theta^{\text{PERL}}\)  
  & Combined parameter vector \((\theta^{\text{Phy}},\theta^{\text{RL}})\) for PERL. \\[6pt]

\(f^{\text{PERL}}(s\mid\theta^{\text{PERL}})\)  
  & PERL predictor: $f^{\text{Phy}}\bigl(S_{\text{Phy}}(s)\mid\theta^{\text{Phy}}\bigr)+f^{\text{RL}}\left(s\mid\theta^{\text{RL}}\right)$. \\[6pt]

\(g(s)\)  
  & Ground‐truth function at state \(s\). \\[6pt]

\(r(s)\)  
  & True residual: \(r(s)=g(s)-f^{\text{Phy}}\bigl(S_{\text{Phy}}(s)\mid\theta^{\text{Phy}}\bigr)\). \\[6pt]

\(L_g\)  
  & Lipschitz constant of \(g(s)\). \\[6pt]

\(L_r\)  
  & Lipschitz constant of the residual \(r(s)\). \\[6pt]

\(\ell^{\rm NN}(\theta^{\rm NN})\)  
  & Expected MSE for pure NN: 
    \(\E_{s\sim\mathcal D}\bigl[f^{\rm NN}(s\mid\theta^{\rm NN})-g(s)\bigr]^2\). \\[6pt]

\(\ell^{\rm RL}(\theta^{\rm RL})\)  
  & Expected MSE for residual learner: 
    \(\E_{s\sim\mathcal D}\bigl[f^{\rm RL}(s\mid\theta^{\rm RL})-r(s)\bigr]^2\). \\[6pt]

\(c_g,\;c_r\)  
  & Uniform upper bounds on \(\ell^{\rm NN}\) and \(\ell^{\rm RL}\), respectively. \\[6pt]

\(\mathcal{E}(L;\eta,T)\)  
  & Convergence‐error bound after \(T\) steps of GD with stepsize \(\eta\) on an \(L\)-Lipschitz loss. \\[6pt]

\(n\)  
  & Sample size (number of i.i.d.\ training points). \\[6pt]

\(\mathcal{D}\)  
  & Data distribution over \((s,y)\). \\[6pt]

\(B\)  
  & Domain‐radius: \(\max_{x\in\Omega}\|x-x^*\|\le B\). \\[6pt]

\(R(f)\), \(\widehat R_n(f)\)  
  & Expected risk \(\E_{\,(s,y)\sim\mathcal D}[\ell(y,f(s))]\) and empirical risk \(\frac1n\sum_i\ell(y_i,f(x_i))\). \\[6pt]

\(\widehat f\), \(f^*\)  
  & Empirical and population risk minimizers over \(\mathcal{F}\). \\[6pt]

\(C\)  
  & Uniform bound: \(\lvert f(s)\rvert,\lvert g(s)\rvert\le C\). \\[6pt]

\(\mathfrak{R}_n(\mathcal{F})\)  
  & Empirical Rademacher complexity of \(\mathcal{F}\) on \(n\) samples. \\
\end{longtable}
\end{center}


This subsection presents the mathematical formulation of the PERL framework~\citep{long2023physics}, including two key assumptions and a set of notations used throughout the analysis. For reader convenience, Table \ref{tab:Notations} summarizes the notations used in this section. To formally characterize the structure of PERL and enable theoretical investigation, we begin by presenting the PERL framework. PERL decomposes the predictions into two components: (i) a physics model $f^{\text{Phy}}(S^{\text{Phy}}(s) \mid \theta^{\text{Phy}})$ that captures the dominant trend of the ground truth function $g(s)$, and (ii) a data-driven NN
model $\displaystyle f^{\text{RL}}(s\mid\theta^{\text{RL}})$ 
that learns the residual 
$r(s)=g(s)-f^{\text{Phy}}\bigl(S^{\text{Phy}}(s)\mid\theta^{\text{Phy}}\bigr),
$
i.e., the difference between the physics model’s output and the ground truth.
\begin{IEEEeqnarray}{rCl}
f^{\text{RL}}\!\bigl(s\mid\theta^{\text{RL}}\bigr)
&=& f^{\text{Phy}}\!\bigl(S^{\text{Phy}}(s)\mid\theta^{\text{Phy}}\bigr)
\label{perl def}\nonumber\\
&& {}+\,f^{\text{RL}}\!\bigl(s\mid\theta^{\text{RL}}\bigr),
\qquad \forall\, s\in\Omega.
\end{IEEEeqnarray}

Without loss of generality, we assume that this framework satisfies the following two properties. 
\subsection{Lipschitz constant reduction}

A function \(f:\Omega\to\mathbb{R}\) is said to be \emph{Lipschitz continuous} if there exists a constant \(L\ge0\) such that
\begin{equation}
  \lvert f(s) - f(s')\rvert \;\le\; L\,\lVert s - s'\rVert,
  \quad\forall\,s,s'\in\Omega.
\end{equation}
We assume the ground‐truth function \(g\) satisfies this with constant \(L_g>0\):
\begin{equation}
\label{eq:lipschitz}
  \lvert g(s) - g(s')\rvert \;\le\; L_g\,\lVert s - s'\rVert,
  \quad\forall\,s,s'\in\Omega.
\end{equation}

From Eq.~\eqref{perl def}, the residual is defined as
$
  r(s) \;=\; g(s)
  \;-\;
f^{\text{Phy}}\bigl(S^{\text{Phy}}(s)\mid\theta^{\text{Phy}}\bigr).
$ Since the physics model captures the dominant behavior of \(g(s)\), the residual \(r(s)\) varies more smoothly and exhibits lower variability. Motivated by this observation, we therefore assume the \emph{Lipschitz Constant Reduction} property: there exists \(L_{r}<L_g\) such that
\begin{equation}
\label{eq:lipschitzReduction}
  \lvert r(s) - r(s')\rvert \;\le\; L_{r}\,\lVert s - s'\rVert,
  \quad\forall\,s,s'\in\Omega.
\end{equation}

\subsection{Training error bound}

Now, consider the standard mean squared error (MSE) loss in a supervised learning setting, where the model \( f^{NN}(\cdot \mid \theta^{NN}) \) aims to directly predict the ground-truth function \( g(\cdot) \) using NN. For an input \( s \in \mathcal{S} \), the prediction loss is defined as:
\begin{equation}
\label{Training loss for NN}
\ell^{\mathrm{NN}}(\theta^{\mathrm{NN}})
=\;\mathbb{E}_{s}\Bigl[\bigl(f^{\mathrm{NN}}(s;\,\theta^{\mathrm{NN}})\;-\;g(s)\bigr)^2\Bigr]\,.
\end{equation}

Let \(f^{\text{RL}}(s \mid \theta^{\text{RL}}) \) denote an approximation of the residual function 
\begin{equation}
\label{eq:residual_function}
r(s)=g(s)-f^{\mathrm{Phy}}\bigl(S^{\mathrm{Phy}}(s)\mid\theta^{\mathrm{Phy}}\bigr),
\end{equation}
the associated MSE loss becomes:
\begin{equation}
\label{Training loss for RL}
\ell^{\mathrm{RL}}(\theta^{\mathrm{RL}})
=\;\mathbb{E}_{s}\Bigl[\bigl(f^{\mathrm{RL}}(s;\theta^{\mathrm{RL}}) - r(s)\bigr)^2\Bigr].
\end{equation}
Intuitively, if the physics model $f^{\text{Phy}}(S^{\text{Phy}}(s) \mid \theta^{\text{Phy}})$ can effectively capture the overall trend of the ground truth function $g(s)$, the resulting residual $r(s)$ will exhibit much lower variability. Therefore, learning the residual function becomes an easier task for the NN, as it only needs to approximate a smoother, lower magnitude target function. This reduced complexity often translates into smaller training error bound. 

Hence we assume the \emph{Training Error Bound Reduction} property: there exists the training loss bound $c_{g}$ and $c_{r}$ for equation \ref{Training loss for NN} and \ref{Training loss for RL} and satisfies $c_{r} < c_{g}.$




\section{Theoretical Proof of PERL}
\label{Theory}
\begin{figure}[H]
    \centering
    \includegraphics[width=0.95\textwidth]{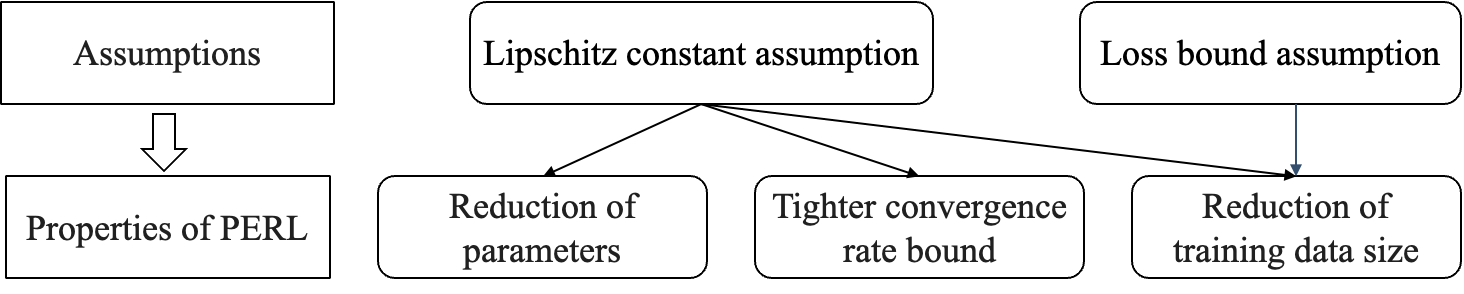}
    \caption{Conceptual framework diagram}
    \label{fig:conceptual_framework_diagram}
\end{figure}

We will use those two properties in section \ref{Review} to theoretically prove that PERL yields the following three benefits compared with pure NNs as shown in figure \ref{fig:conceptual_framework_diagram}:
\begin{itemize}
    \item[(1)] A reduction in the number of parameters in NNs architecture.
    \item[(2)] A tighter bound on the convergence rate during training.
    \item[(3)] A reduction in the required training data size to achieve a specified bound on both generalization and estimation errors.
\end{itemize}

\subsection{The number of parameters}
In a NN, the number of parameters is the sum of the number of weights and biases. The parameters are represented as a linear transformation within each neuron, and in the hidden layers, those parameters are trained through back propagation. According to the Universal Approximation Theorem (UAT), a shallow NN has the capacity to approximate any continuous function \citep{cybenko1989approximation}. Therefore, in this analysis, we consider a two-layer NN with the ReLU activation function, which apply element-wise nonlinearity to the neuron outputs, resulting in either zero or a linear function segment. This structure enables the network to approximate any continuous function using piecewise linear segments \citep{hornik1991approximation}. The number of required segments correlates with the number of parameters needed in training, and more segments imply a higher parameter requirement. We formally present the following theorem and the proof is in Appendix \ref{pf: number of parameters}:

\begin{theorem}
\label{theorem: number of parameters}

Assume \(a,b\in\mathbb{R}\) and let 
$\mathcal{F}\subset\{\,f:[a,b]\to\mathbb{R}\}$
be a family of functions defined on the closed interval \([a,b]\), with each \(f\in\mathcal{F}\) being Lipschitz continuous with the same Lipschitz constant \(L\).  Let each function \(f\in\mathcal{F}\) be approximated by a piecewise linear function \(\hat{f}(x)\) over the interval \([a,b]\), such that the total approximation error does not exceed a given tolerance \(\varepsilon>0\):

\begin{equation}
\int_a^b |f(x) - \hat{f}(x)| \, dx \leq \varepsilon.
\end{equation}
Then, to achieve this level of accuracy for every \( f \in \mathcal{F} \), the supremum on the minimum number of linear segments required in the piecewise linear approximation is
\begin{equation}
P = \left\lceil \dfrac{L(b - a)^2}{4 \varepsilon} \right\rceil.
\end{equation}
Here, \(\lceil \cdot \rceil\) is the ceiling function, which rounds the value up to the nearest integer.
\end{theorem}

Theorem \ref{theorem: number of parameters} establishes a relationship between the number of required linear segments \( N \) and the Lipschitz constant \( L \) in NNs. Specifically, it shows that \( N \) is proportional to \( L \), meaning that as \( L \) decreases, the required \( N \) also decreases. Based on this theorem, we have the following proposition.


\begin{proposition}
Let \(f\) be \(L_f\)-Lipschitz continuous, and define the residual function 
\(r(s) = g(s) - f^{\mathrm{Phy}}\bigl(S^{\mathrm{Phy}}(s)\mid\theta^{\mathrm{Phy}}\bigr)\),
which is \(L_r\)-Lipschitz continuous with \(L_r < L_f\). For any \(\varepsilon>0\), let
$P_f(\varepsilon),\,P_r(\varepsilon)$
denote the minimal number of parameters of a two‐layer neural network required to achieve
the same approximation accuracy \(\varepsilon>0\) of function $g$ and $r$,
respectively.  Then
\begin{equation}
P_r(\varepsilon)\;<\;P_f(\varepsilon),
\end{equation}
i.e.\ achieving the same error \(\varepsilon\) requires fewer parameters when learning the residual $r$ than when learning \(f\) directly.
\end{proposition}

This proposition is also intuitive: when a NN is tasked with predicting a smoother function, it generally requires fewer parameters compared to predicting a highly changeable function. After preprocessing with a physics model to remove the dominant behavior, the residual learning network does not need to allocate resources to model large-scale fluctuations, allowing for a reduction in the number of neurons and layers. Consequently, the NN size can be reduced without compromising performance.

Although our formal bound focuses on one-dimensional inputs, the same idea extends to higher dimensions.  Multilayer ReLU networks partition \(\mathbb{R}^d\) into piecewise-linear regions, and since a lower Lipschitz constant requires fewer such regions to achieve a uniform approximation error \(\varepsilon\), fewer hidden units and hence fewer parameters are needed when \(d>1\).  Consequently, even in higher-dimensional settings, applying our residual-learning theorem after physics preprocessing yields a strict reduction in network size.

\subsubsection{Example}
\label{example}
\begin{figure}[ht!]
    \centering
    \includegraphics[width=0.7\textwidth]{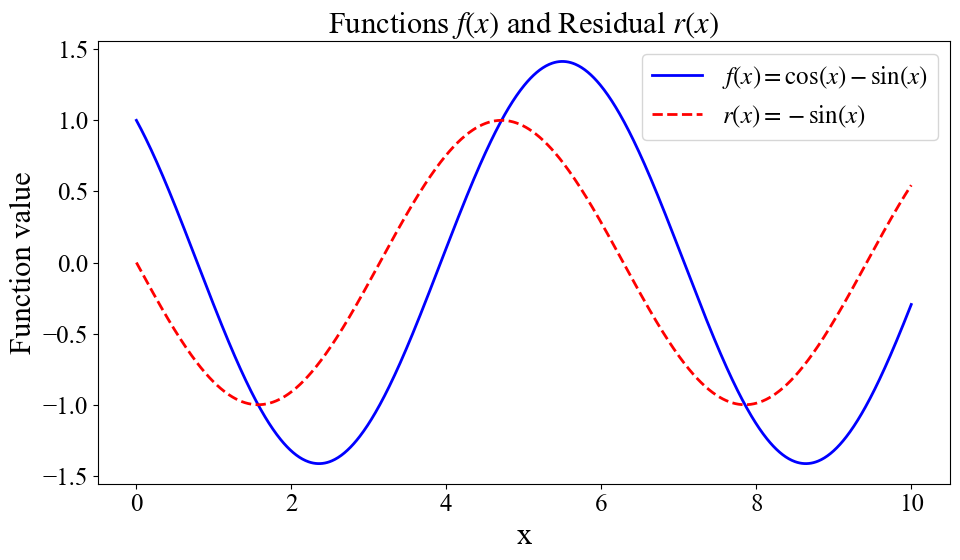} 
    \caption{Illustration of the original function f(x) and the residual function r(x).}
    \label{fig:example_image}
\end{figure}
Consider the target function \( f(x) = \cos(x) - \sin(x) \) defined on the closed interval \([0, 10]\). Suppose we use the physical model prediction \( \hat{f}(x) = \cos(x) \), resulting in the residual function \( r(x) = f(x) - \hat{f}(x) = -\sin(x) \), as illustrated in Figure \ref{fig:example_image}. The Lipschitz constant of the target and residual functions over this interval are \( L_f = \sqrt{2} \) and \( L_r = 1 \), respectively.

We use piecewise linear functions to approximate \( f(x) \) and \( r(x) \). For the same tolerance \( \epsilon \), we determine the number of pieces required for approximating each function. As we vary the value of \( \epsilon \), we record the results, which are presented in Table 
\ref{tab:pieces}. The residual function \( r(x) \) with a smaller Lipschitz constant consistently requires fewer pieces than the original target function \( f(x) \), with an average reduction of 15\%. This example demonstrates that for a shallow NN with ReLU activation function, the number of parameters required by PERL is reduced compared to traditional NNs.



\begin{table}[ht]
    \caption{Number of pieces required for approximating $f(x)$ and $r(x)$ at different $\epsilon$ values.}
    \centering
    \begin{tabular}{c|c|c|c}
        \hline
        $\epsilon\ (\times 10^{-3})$ & \# pieces for $f(x)$ & \# pieces for $r(x)$ & Reduction (\%) \\
        \hline
        1  & 134 & 113 & 15.67 \\
        2  & 95  & 81  & 14.74 \\
        3  & 78  & 66  & 15.38 \\
        4  & 68  & 57  & 16.18 \\
        5  & 61  & 51  & 16.39 \\
        6  & 56  & 47  & 16.07 \\
        7  & 52  & 44  & 15.38 \\
        8  & 48  & 41  & 14.58 \\
        9  & 46  & 39  & 15.22 \\
        \hline
    \end{tabular}
    \label{tab:pieces}
\end{table}

\subsection{Convergence rate}
Gradient descent is a fundamental algorithm for training NNs, offering an efficient approach to minimizing the loss function through iterative updates along the direction of steepest descent \citep{du2019gradient}. Analyzing the convergence behavior under gradient descent reveals how key factors such as the step size \( \eta \) and the Lipschitz constant \( L \) influence the training process. This understanding provides practical insights into NN optimization and informs strategies to accelerate convergence.

To compare the convergence rate between the PERL framework and a pure NN, we first consider a special case where the objective functions \( f \in \mathcal{F} \) are convex and $L$-Lipschitz continuous. Consider the general gradient descent iteration given by \citep{bubeck2015convex}
\begin{equation}
\boldsymbol{x}^{t+1} = \boldsymbol{x}^t - \eta_t \nabla f\left(\boldsymbol{x}^t\right),
\end{equation}
where \(\boldsymbol x^t\in\mathbb R^d\) is the parameter vector at iteration \(t\), \(\nabla f(\boldsymbol x^t)\) is its gradient, and \(\eta_t>0\) is the step size.

Starting from an arbitrary initialization \(\boldsymbol x\in\mathbb R^d\), the general solution for this convex optimization problem is given by:
\begin{equation}
\boldsymbol{x}^*_f=\arg \min _{\boldsymbol{x} \in \mathbb{R}^d} f(\boldsymbol{x})
\end{equation}
The convergence rate is bounded by the following theorem \citep{zinkevich2003online}.

\begin{theorem}
\label{theorem: convergence rate bound}
Assume we have a class of convex, differentiable objective functions $\mathcal{F}$, where each function $f \in \mathcal{F}$ is Lipschitz continuous with the same Lipschitz constant $L$ for all $\boldsymbol{x}$. The domain of $\boldsymbol{x}$ is bounded for each $f$ with bound $B$, that is,
\begin{equation}
\max_{\boldsymbol{x}} \left\| \boldsymbol{x} - \boldsymbol{x}_f^* \right\| \leq B, \quad \forall f \in \mathcal{F}
\end{equation}
where $\boldsymbol{x}_f^*$ denotes  a global minimizer of $f$. Each function $f \in \mathcal{F}$ is trained using gradient descent with a constant step size $\eta$ for $T$ iterations. Then, for each $f \in \mathcal{F}$, the average convergence error is bounded as:
{\normalfont
\begin{IEEEeqnarray}{rCl}
\mathcal{E}(L;\eta,T)
&=& \frac{1}{T}\sum_{t=1}^T\!\bigl(f(\boldsymbol{x}^t)-f(\boldsymbol{x}^*_f)\bigr) \nonumber\\
&\le& \frac{B^2}{2\eta T}+\frac{\eta L^2}{2}.
\end{IEEEeqnarray}
}
\end{theorem}
From this convergence error, we can observe that the convergence rate depends on two terms: \( \dfrac{B^2}{2 \eta T} \), and \( \dfrac{\eta L^2}{2} \).  The first term \( \dfrac{B^2}{2 \eta T} \) decreases with more iterations \( T \), and the second term \( \dfrac{\eta L^2}{2} \) decreases with a smaller Lipschitz constant.


\begin{proposition}
\label{prop:gradient descent}
Let $g(s)$ be convex and \(L_g\)-Lipschitz continuous. Under the residual learning framework of PERL, and assume $r(s)$ is $L_{r}$‐Lipschitz with $L_{r}<L_g$.  Let $\mathcal{E}_g \;=\;E(L_g;\eta,T),$ and
$\mathcal{E}_{r} \;=\;E(L_{r};\eta,T),$
be the average convergence error bound after $T$ steps of gradient descent with step‐size $\eta$.  Then
\begin{equation}
\mathcal{E}_{r}\;<\;\mathcal{E}_g,
\end{equation}
i.e.\ under identical $\eta$ and $T$, the residual learning framework in PERL converges faster than a pure NN trained directly.
\end{proposition}

This proposition \ref{prop:gradient descent} gives two insightful conclusions about PERL framework. First, for any fixed step‐size \(\eta>0\) and iteration count \(T\), lowering the Lipschitz constant \(L\) directly tightens the convergence error bound \(E(L;\eta,T)\). Since the PERL residual has \(L_{r}<L_g\), it follows that 
$E(L_{r};\eta,T)<E(L_g;\eta,T),$
so PERL converges faster than a pure NN under identical training settings. Second, in the asymptotic regime (\(T\to\infty\), \(\eta\to0\)), and under the usual smoothness and convexity assumptions, both the standard NN and the PERL framework are guaranteed to converge to their global optima, as predicted by classical gradient descent theory.

In practical training, the constant step size \( \eta \) can be chosen as \( \eta = \dfrac{1}{L} \). Based on our previous discussion and analysis, this choice is justified because it ensures convergence for convex functions with Lipschitz continuous gradients. Substituting \( \eta = \dfrac{1}{L} \) into the convergence bound simplifies it to \citep{nesterov2013introductory}:
\begin{equation}
\dfrac{1}{T} \sum_{t=1}^T \bigl( f\left( \boldsymbol{x}^t \right) - f\left( \boldsymbol{x}_f^* \right) \bigr) \leq \dfrac{B^2 L}{2 T} + \dfrac{L}{2}.
\end{equation}
With this step size, both the term \( \dfrac{B^2 L}{2 T} \) and \( \dfrac{L}{2} \) decrease as the Lipschitz constant \( L \) becomes smaller. However, when the Lipschitz constant $L$ is considerably large, the step size $\dfrac{1}{L}$ will be dramatically small, which may lead to slow convergence. An effective approach to address this issue is to use a diminishing step size, based on the following lemma and theorem. The proofs appear in Appendix \ref{pf:integral} and \ref{pf:non-constant-step-size-convergence-rate-bound}

\begin{lemma}
\label{lemma: integral}
For any \( t = 1, 2, \dots, \) the following inequality holds:
\begin{equation}
\sum_{t=1}^T \dfrac{1}{\sqrt{t}} \leq 2 \sqrt{T}
\end{equation}  
\end{lemma}
Based on the lemma above, we show the theory with diminishing step size.

\begin{corollary}
\label{Corollary: non-constant step size convergence rate bound}
Assume we have a class of convex, differentiable objective functions $\mathcal{F}$, where each function $f \in \mathcal{F}$ is Lipschitz continuous with the same Lipschitz constant $L$ for all $\boldsymbol{x}$. The domain of $\boldsymbol{x}$ is bounded for each $f$ with bound $B$, that is,
\begin{equation}
\max_{\boldsymbol{x}} \left\| \boldsymbol{x} - \boldsymbol{x}_f^* \right\| \leq B, \quad \forall f \in \mathcal{F}
\end{equation}
where $\boldsymbol{x}_f^*$ is the global minimizer of $f$. Each function $f \in \mathcal{F}$ is trained using gradient descent with a diminishing step size $\eta_t = \dfrac{1}{\sqrt{t}}$ for $T$ iterations. Then, for each $f \in \mathcal{F}$, the average convergence rate is bounded as:
{\normalfont
\begin{IEEEeqnarray}{rCl}
\mathcal{E}(L;\eta_t,T)
&=& \frac{1}{T}\sum_{t=1}^T\!\bigl(f(\boldsymbol{x}^t)-f(\boldsymbol{x}^*_f)\bigr) \nonumber\\
&\le& \frac{1}{\sqrt{T}}\!\left(\frac{B^2}{2}+L^2\right).
\end{IEEEeqnarray}
}
\end{corollary}

Corollary \ref{Corollary: non-constant step size convergence rate bound} shows that, under the diminishing step‐size schedule $\eta_t = \dfrac{1}{\sqrt{t}}$, the convergence error bound decays at rate $O(\dfrac{1}{\sqrt T})$.  In this form, both the domain bound term $\dfrac{B^2}{2}$ and the Lipschitz term $L^2$ vanish as $T\to\infty$, unlike the constant‐step case where an $O(1)$ term remains.  This sublinear $O(\dfrac{1}{\sqrt T})$ rate is known to be optimal for general convex minimization under first‐order methods \citep{bubeck2015convex}, and it obviates the need for excessively small fixed step sizes when $L$ is large that would otherwise slow down convergence.


Although we assume convex \(L\)-Lipschitz functions throughout this subsection to guarantee a unique minimizer \(\boldsymbol x^*_f\), our comparison depends only on the objective value gap $f(\boldsymbol x^t) - f(\boldsymbol x^*)$ as a measure of convergence. As a result, strict convexity or uniqueness of \(\boldsymbol x^*\) is not required since tracking the decrease in loss alone suffices. In the PERL framework, physics model preprocessing produces a residual function with Lipschitz constant \(L_{r} < L_g\), so the bound $\mathcal{E}_{r}\;<\;\mathcal{E}_g$ still holds. Therefore, the faster convergence of the PERL framework is driven solely by the reduced Lipschitz constant and does not depend on strict convexity. We now illustrate this accelerated convergence in a simple numerical example using constant step size \(\eta = \dfrac{1}{10L}\).

\subsubsection{Example}
\begin{figure}[htbp]
    \centering
    \includegraphics[width=0.7\textwidth]{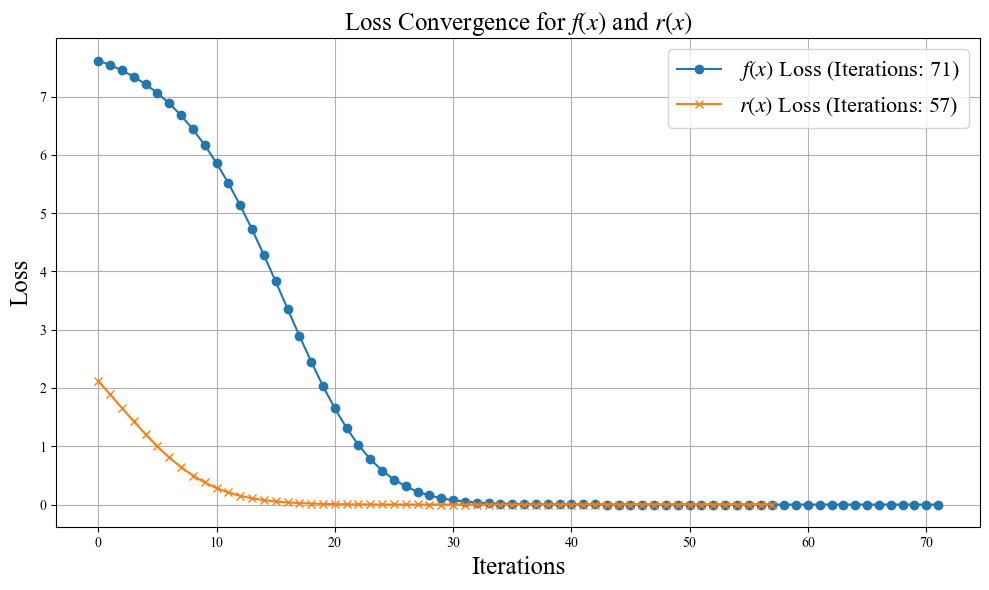}
    \caption{Number of iterations required for f(x) and r(x) to converge with the given tolerance value of 0.001.}
    \label{fig:convergence}
\end{figure}
Using the same example functions as in Section~\ref{example}, 
where \(f(x) = \cos x - \sin x\) and \(r(x) = -\sin x\), 
with Lipschitz constants \(L_g = \sqrt{2}\) and \(L_r = 1\), we set the tolerance to \( 10^{-3} \). With a constant update step size of \( \dfrac{1}{10L} \), \( f(x) \) requires 71 iterations to converge, whereas \( r(x) \) requires only 57 iterations to converge, as illustrated in Figure 
\ref{fig:convergence}. 

By running gradient descent on both $f(x)$ and $r(x)$ under the same conditions, this example confirms the conclusions in this subsection. It shows that after physics model preprocessed, the reduced Lipschitz constant tightens the convergence error bound and produces faster convergence in practice. Moreover, the residual learning completes training in 20\% fewer iterations, demonstrating the benefit of bound reduction. Together, these findings illustrate the effectiveness of the PERL framework.

\subsection{Error bound}
In statistical learning theory, evaluating a model's performance involves how well its predictions align with the true labels. This is typically done by analyzing the expected risk and the empirical risk \citep{bartlett2002rademacher} which are defined as follows:

\begin{definition}
\label{expected risk}
For a function \( f : \mathbb{R}^d \to \mathbb{R} \) belonging to a function class \(\mathcal{F}\) and a loss function \(\ell : \mathbb{R} \times \mathbb{R} \to \mathbb{R}_+\), the expected risk is defined as
\begin{equation}
R(f) = \mathbb{E}_{(x,y) \sim D} \left[\ell\bigl(y, f(x)\bigr)\right],
\end{equation}
where \( D \) is the underlying probability distribution,
\end{definition}

\begin{definition}
\label{empirical risk}
Given a training sample set \( S = \{(x_1, y_1), (x_2, y_2), \ldots, (x_n, y_n)\} \) drawn independently from the distribution \( D \), the empirical risk is defined as
\begin{equation}
\hat{R}_n(f) = \dfrac{1}{n} \sum_{i=1}^{n} \ell\bigl(y_i, f(x_i)\bigr).
\end{equation}

\end{definition}

\subsubsection{Estimation error}

The estimation error quantifies the discrepancy between the performance of the model learned from finite data and the best possible model within the function class. Formally, let
\begin{equation}
f^\star = \arg\min_{f \in \mathcal{F}} R(f)
\end{equation}
denote the optimal predictor with respect to the true distribution \(D\), and let
\begin{equation}
\hat{f} = \arg\min_{f \in \mathcal{F}} \hat{R}_n(f)
\end{equation}
be the predictor obtained by minimizing the empirical risk based on a training sample of size \(n\). Then, we can define the estimation error as follows:
\begin{definition}
\begin{equation}
R(\hat{f}) - R(f^\star).
\end{equation}
\end{definition}
This error measures how much the risk of the learned model exceeds the optimal risk in \(\mathcal{F}\), reflecting the effect of using a limited sample for parameter estimation. In essence, it captures the uncertainty and variance inherent in the learning process due to finite data. To establish the estimation error bound, we first introduce the Hoeffding inequality as shown in the following lemma \citep{hoeffding1994probability}.
\begin{lemma}[Hoeffding Inequality]
\label{lemma: Hoeffding Inequality}
Let \(X_1,\dots,X_n\) be independent sub‑Gaussian random variables with variance proxy \(\sigma^2\); that is, for every real \(s\) and each \(i\),
\begin{equation}
\mathbb{E}\bigl[e^{s\,(X_i - \mathbb{E}[X_i])}\bigr]
\;\le\;\exp\!\Bigl(\frac{s^2\,\sigma^2}{2}\Bigr).
\end{equation}

Define \(S_n=\sum_{i=1}^nX_i\).  Then for any \(t>0\),
\begin{equation}
    \Pr\bigl(|S_n-\mathbb{E}[S_n]|\ge t\bigr)
\;\le\;2\exp\!\Bigl(-\frac{t^2}{2n\,\sigma^2}\Bigr).
\end{equation}
\end{lemma}
According to Lemma~\ref{lemma: Hoeffding Inequality}, any sequence of independent sub‑Gaussian random variables has an exponential tail bound on its deviation from the mean. Let \(\{(x_i,y_i)\}_{i=1}^n\) be drawn i.i.d.\ from the data distribution $D$.  Under Assumption~2, define the sample loss
$\ell_i = \ell\bigl(f, x_i, y_i\bigr),$
which satisfies $\ell_i \in [0,\,c_f].$
It follows from Hoeffding’s lemma that any random variable lied on an interval of length \(c_f\) is sub‑Gaussian with variance proxy \(\sigma^2 = \dfrac{c_f^2}{4}\) \citep{hoeffding1963probability}.  In the statement of Lemma~\ref{lemma: Hoeffding Inequality}, the sub‑Gaussian parameter \(\sigma^2\) is exactly \((b-a)^2/4\) when \(X_i\in[a,b]\).  Therefore, setting \(\sigma^2 = c_f^2/4\) and applying Lemma~\ref{lemma: Hoeffding Inequality} with \(X_i = \ell(f,x_i,y_i)\) yields  


\begin{equation}
\mathbb{P}(|\hat{R}(f) - R(f)| > t) \;\le\;
2\exp\!\Bigl(-\frac{2\,n\,t^2}{c_f^2}\Bigr).\end{equation}

Before deriving the bound for estimation error, we obtain the following lemma, and the proof is in Appendix \ref{pf: Inequality}.

\begin{lemma}
\label{Inequalities}
The following inequalities hold with probability at least $1 - \delta$:
\begin{equation}
R(\hat{f}) \leq \hat{R}(\hat{f}) + t
\leq \hat{R}(f^\star) + t
\leq R(f^\star) + 2t,
\end{equation}
where $\delta = 2\exp\!\Bigl(-\dfrac{2nt^2}{c_f^2}\Bigr)$.
\end{lemma}
Based on the Lemma \ref{lemma: Hoeffding Inequality} and Lemma \ref{Inequalities} above, we show the theorem that bound the estimation error and the proof is summarized in Appendix \ref{pf: estimation error}.

\begin{theorem}
\label{theorem: estimation error}
For any sample size \( n \) and any \( \epsilon > 0 \), given a function class \( \mathcal{F} \), the probability of estimation error is bounded by the following inequality:
\begin{equation}
\mathbb{P}(R(\hat{f}) - R(f^*) \geq \epsilon) \leq 4\exp\!\Bigl(-\frac{n\epsilon^2}{2c_f^2}\Bigr),
\end{equation}
 where \( c_f \) is a upper bound to the function $f$.
\end{theorem}
    
According to the inequality above, the probability that the estimation error exceeds \(\epsilon\) is bounded by $4\exp\!\Bigl(-\dfrac{n\epsilon^2}{2c_f^2}\Bigr)$. This bound reveals that, as the sample size \( n \) increases, the probability that the estimation error exceeds a fixed threshold \( \epsilon \) decreases exponentially.

\begin{proposition}
Let \(\varepsilon>0\) and \(\delta\in(0,1)\).  Under the conditions of Theorem \ref{theorem: estimation error}, any function $f$ whose sample loss is bounded in \([0,c_f]\) requires a sample size
\begin{equation}
n \;\ge\;\frac{c_f^2}{2\varepsilon^2}\,\ln\!\Bigl(\frac{4}{\delta}\Bigr)
\end{equation}
to ensure
\(\mathbb{P}\bigl(|\widehat R(f)-R(f)|\le\varepsilon\bigr)\ge1-\delta\).  

Now according to the Assumption 2, let \(g(s)\) and \(r(s)\) be functions whose sample losses satisfy
$
\ell\bigl(g(s),y\bigr)\in[0,c_g],$
and
$
\ell\bigl(r(s),y\bigr)\in[0,c_r],
$
with \(0\le c_r<c_g\).  Define
\begin{equation}
  N_g = \frac{c_g^2}{2\varepsilon^2}\,\ln\!\Bigl(\frac{4}{\delta}\Bigr),
\quad
N_r = \frac{c_r^2}{2\varepsilon^2}\,\ln\!\Bigl(\frac{4}{\delta}\Bigr).  
\end{equation}
Then
\begin{equation}
N_r \;<\; N_g,
\end{equation}
i.e.\ the lower bound on the required sample size for achieving the same estimation‑error level \(\varepsilon\) at confidence \(1-\delta\) is strictly smaller when training the residual function \(r(s)\) in PERL than when training the ground truth function \(g(s)\) in pure NN.

\end{proposition}

This proposition highlights one of the key theoretical advantages of the PERL framework: by leveraging a physics model to approximate the dominant trend of the prediction, the residual function $r(s)$ requires a smaller sample size \( n \) to achieve the same estimation accuracy given the smaller loss bound. On the contrary, pure NNs require more training data to reach comparable accuracy.

\subsubsection{Generalization Error}

The generalization error quantifies the discrepancy between the model's performance on the true distribution and its performance on the training data, thereby reflecting the model's ability to generalize to new, unseen data. Based on the definition \ref{expected risk} and \ref{empirical risk}, the generalization error is defined as

\begin{definition}
\begin{equation}
R(f) - \hat{R}_n(f).
\end{equation}
\end{definition}

To bound the generalization error, we introduce the concept of Rademacher Complexity, a widely used measure of the complexity of a function class.

\begin{definition}
For a function class \(\mathcal{F}\) and a sample \(S = \{x_1, x_2, \ldots, x_n\}\) drawn from a distribution \(D\), the Rademacher Complexity is defined as
\begin{equation}
\mathcal{R}_n(\mathcal{F}) := \mathbb{E}_{\sigma}\left[\sup_{f \in \mathcal{F}} \dfrac{1}{n} \sum_{i=1}^{n} \sigma_i f(x_i)\right],
\end{equation}
where \(\sigma_1, \sigma_2, \ldots, \sigma_n\) are independent Rademacher random variables with
$\mathbb{P}(\sigma_i = 1) = \mathbb{P}(\sigma_i = -1) = \dfrac{1}{2}.$
\end{definition}
Based on the definition above, the generalization error can be bounded by the Rademacher Complexity, the Lipschitz constant, and the sample size \citep{bartlett2002rademacher}.
\begin{theorem}
\label{theorem:generalization}
Let $\mathcal{F}$ be a class of functions, $\{(x_i, y_i)\}_{i=1}^n$ be iid training examples, and $\ell$ be an $L$-Lipschitz loss function. Consider the empirical risk function $\hat{R}_n(f) $ and its expectation $R(f)$. Assume the losses are bounded in $[0, c]$. With probability at least $1 - \delta$
\begin{equation}
 R(f) - \hat{R}_n(f) \leq 2 \mathcal{R}_n(\mathcal{\ell (F)}) + c \sqrt{\dfrac{\log(1/\delta)}{2n}},
\end{equation}
where
$$
\mathcal{R}_n(\mathcal{\ell(F)}) = \mathbb{E}\left[\sup_{f \in \mathcal{F}} \dfrac{1}{n} \sum_{i=1}^{n} \sigma_i \ell_i(f(x_i))\right].
$$
\end{theorem}

To further analyze the generalization bound in Theorem~\ref{theorem:generalization}, it is crucial to disentangle the complexity contribution of the loss function $\ell$ from that of the function class $\mathcal{F}$. Since the Rademacher complexity involves the composed class $\ell \circ \mathcal{F}$, directly bounding this can be difficult. However, when $\ell$ is Lipschitz continuous, we can leverage Lemma~\ref{lemma:contraction} to control the complexity of $\ell \circ \mathcal{F}$ by that of $\mathcal{F}$ itself, scaled by the Lipschitz constant. In our setting, since $f \in \mathcal{F}$ is assumed to be $L$-Lipschitz and defined on a compact domain, $f(x)$ is bounded according to Weierstrass extreme value theorem. Combined with the fact that we use the MSE as the loss function, it follows that the loss function $\ell$ is also Lipschitz continuous, with a constant denoted by $L_\ell$. We prove this statement in the following lemma and show the relationship between $L_\ell$ and $L$.

\begin{lemma}
\label{lemma: Loss Lip constant}
Suppose the function \( f \) is \(L\)-Lipschitz continuous with respect to \(s\), and that both \(f(s)\) and the ground truth function \(g(s)\) are bounded by a constant \(C\), i.e., \( |f(s)| \leq C \) and \( |g(s)| \leq C \). Then the MSE loss function
\begin{equation}
\ell(s;\theta) \;=\;\bigl(f(s;\,\theta) - g(s)\bigr)^2.
\end{equation}
is Lipschitz continuous with respect to \(s\), with Lipschitz constant \(L_\ell = 4CL\).
\end{lemma}
Lemma \ref{lemma: Loss Lip constant} ensures that the loss function $\ell$ satisfies the Lipschitz condition required to apply the following lemma \ref{lemma:contraction}.  A proof of this result can be found in \citet{ledoux2013probability}.
\begin{lemma}
\label{lemma:contraction}
Suppose $\{\phi_i\}$ and $\{\psi_i\}$ are two sets of functions on domain $\mathcal{F}$ such that for each $i$ and $f, f' \in \mathcal{F}$,
\begin{equation}
|\phi_i(f) - \phi_i(f')| \leq |\psi_i(f) - \psi_i(f')|.
\end{equation}
Then
\begin{equation}
\mathbb{E}_{\sigma}\left[\sup_f \sum_{i=1}^{n} \sigma_i \phi_i(f)\right] \leq \mathbb{E}_{\sigma}\left[\sup_f \sum_{i=1}^{n} \sigma_i \psi_i(f)\right].
\end{equation}
\end{lemma} 
The lemma \ref{lemma:contraction} provides a general inequality that allows us to bound the Rademacher complexity of a set of transformed functions using the complexity of the original functions. Specifically, the transformation $f \mapsto \ell(f)$ is $L_\ell$-Lipschitz, which allows us to bound the Rademacher complexity of the loss function class $\ell \circ \mathcal{F}$ in terms of that of $\mathcal{F}$. Apply the lemma above with $\phi_i(f) = \varphi \left( z_i(f) \right), \ \psi_i(f) = L z_i(f)$. This leads to the following lemma:

\begin{lemma}
\label{lemma:contraction_loss}
Consider a finite collection of stochastic processes $z_1(f), z_2(f),$ $ \ldots, z_n(f)$ indexed by $f \in \mathcal{F}$. Let $\sigma_1, \ldots, \sigma_n$ be independent Rademacher random variables. Then for any $L_\ell$–Lipschitz function $\ell$
\begin{equation}
\mathbb{E} \left[ \sup_{f \in \mathcal{F}} \sum_{i=1}^{n} \sigma_i \ell(z_i(f)) \right] \leq L_\ell \, \mathbb{E} \left[ \sup_{f \in \mathcal{F}} \sum_{i=1}^{n} \sigma_i z_i(f) \right].
\end{equation}
\end{lemma}

Combine these lemmas \ref{lemma: Loss Lip constant}, \ref{lemma:contraction}, \ref{lemma:contraction_loss}, we establish a bound on the Rademacher complexity of the composed loss function class \( \ell \circ \mathcal{F} \) in the following theorem and the proof is in Appendix \ref{pf: generalization error}.

\begin{theorem}
\label{thm:generalization_w/o loss}
Let $\mathcal{F}$ be a class of $L$-Lipschitz functions \(f : \Omega \to \mathbb{R}\), defined on a compact domain, and let \(g(s)\) be the ground-truth function such that both \(f(s)\) and \(g(s)\) are bounded by a constant \(C\). Let the loss function be the MSE: $\ell(s;\theta) \;=\;\bigl(f(s;\,\theta) - g(s)\bigr)^2.
$ Let \(\mathcal{R}_n(\mathcal{F})\) denote the empirical Rademacher complexity of the function class \(\mathcal{F}\), and let \(n\) be the number of training samples. Assume further that the loss \(\ell(s; \theta)\) is bounded in \([0, c]\) for some constant \(c\). Then, with probability at least \(1 - \delta\), the generalization error satisfies:
\begin{equation}
R(f) - \hat{R}_n(f) \leq 8CL \mathcal{R}_n(\mathcal{F}) + c \sqrt{\dfrac{\log(1/\delta)}{2n}},
\end{equation}
\end{theorem}

The inequality shown establishes an upper bound on the discrepancy between the expected risk \( R(f) \) and the empirical risk \( \hat{R}_n(f) \) over a function class \(\mathcal{F}\). Specifically, the Rademacher complexity term \( \mathcal{R}_n(\ell \circ \mathcal{F}) \) can be controlled by scaling the Rademacher complexity of \(\mathcal{F}\) by a factor of \(8CL\). In this generalization bound, the term \(8CL \mathcal{R}_n(\mathcal{F})\) depends directly on the Lipschitz constant \(L\). Lowering \(L\) reduces this bound, thereby tightening the difference between \( R(f) \) and \( \hat{R}_n(f) \) with probability at least 1 - $\delta$. Additionally, a smaller constant \(c\) in the second term, \(c \sqrt{\dfrac{\log(1/\delta)}{2n}}\), further tightens the bound. Consequently, for a fixed generalization error bound, a continuous function with smaller constants \(c\) and \(L\) requires a smaller sample size \(n\) to achieve the same level of generalization. This leads to the following proposition.

\begin{proposition}
\label{prop:perl_sample_complexity}
Under the conditions of Theorem~5, fix any \(\varepsilon>0\) and \(\delta\in(0,1)\).  Let
\begin{equation}
n_g = \min\bigl\{n\in\mathbb{N}:\mathbb{P}(R(f)-\widehat R_n(f)\le\varepsilon)\ge1-\delta\bigr\},
\end{equation}
where ground truth function \(g\) is \(L_g\)-Lipschitz with sample loss bounded in \([0,c_g]\).  Likewise, let
\begin{equation}
n_r = \min\bigl\{n\in\mathbb{N}:\mathbb{P}(R(r)-\widehat R_n(r)\le\varepsilon)\ge1-\delta\bigr\},
\end{equation}
for the PERL residual function \(r\) with effective constants \(L_r\) and \(c_r\) satisfying \(L_r < L_g\) and \(c_r < c_g\).
Then
\begin{equation}
n_r < n_g,
\end{equation}
i.e.\ the lower bound on the required sample size for achieving the same generalization error level \(\varepsilon\) at confidence \(1-\delta\) is smaller when training the residual function \(r(s)\) in PERL than when training the ground truth function \(g(s)\) in pure NN.
\end{proposition}

This proposition highlights the theoretical advantage of the PERL framework. By leveraging a physics model to approximate the dominant trend of the target function, PERL effectively reduces both the Lipschitz constant and the bounded loss in the learning process. As a result, the residual function $r(s)$ becomes smoother and more predictable, which, according to the generalization bound established earlier, requires fewer training samples to achieve the same level of generalization error. This aligns with our intuitive analysis that PERL can achieve better accuracy under limited data regimes. The PERL framework thus offers a reliable and sample-efficient alternative to purely black-box NN approaches.

\subsubsection{Example}
\begin{figure}[htbp]
  \centering
  \includegraphics[width=1\textwidth]{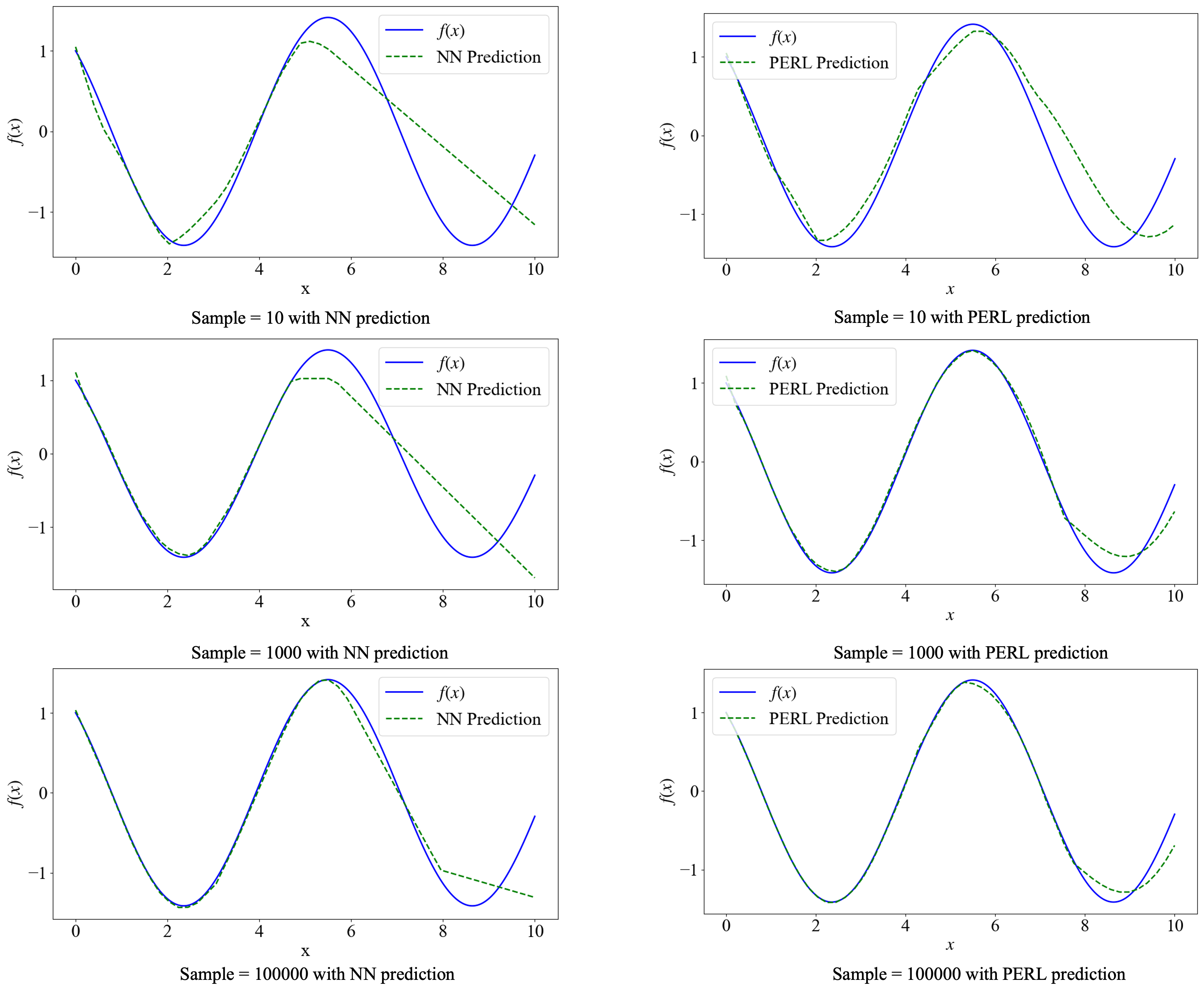}
  \caption{Comparison of predictions from the PERL framework versus a pure NN across different sample sizes.}
  \label{fig:comparison_samples}
\end{figure}

We use the same example functions as in Section~\ref{example}, 
where \(f(x) = \cos x - \sin x\) and \(r(x) = -\sin x\), 
with Lipschitz constants \(L_g = \sqrt{2}\) and \(L_r = 1\).
Using two-layer multilayer perceptrons for both PERL and the pure NN to fit these functions, with 128 and 64 neurons in each layer respectively. As we vary the number of samples, the fitting results are shown in Figure \ref{fig:comparison_samples}. The estimation error and generalization error are presented in Table \ref{tab:error_info}. It can be seen that for both error metrics, PERL exhibits significant improvements compared to NN, especially when the sample size is relatively small. Moreover, since real‐world datasets often have limited sample sizes, our PERL framework’s ability to achieve high accuracy with few observations makes it especially effective in practice.
Besides, PERL’s ability to incorporate physical priors also allows it to maintain robust predictive performance under severe data scarcity.

\begin{table}[htbp]
\centering
\caption{Comparison of $f(x)$, $r(x)$, Generalization Error, and Estimation Error for Different Sample Sizes}
\begin{tabular}{cccccc}
\toprule
Sample Size & Function & Gen. Error & Est. Error & Gen. Improvement (\%) & Est. Improvement (\%) \\
\midrule
10   & \(f(x)\) & 0.1640 & 0.1984 & –     & –     \\
10   & \(r(x)\) & 0.1126 & 0.1114 & 31.3  & 43.9  \\
\midrule
1000 & \(f(x)\) & 0.1565 & 0.1316 & –     & –     \\
1000 & \(r(x)\) & 0.0344 & 0.0342 & 78.0  & 74.0  \\
\midrule
100000 & \(f(x)\) & 0.0602 & 0.0477 & –     & –     \\
100000 & \(r(x)\) & 0.0156 & 0.0145 & 74.1  & 69.6  \\
\bottomrule
\end{tabular}
\label{tab:error_info}
\end{table}

\section{Experiments}
\label{experiments}
\subsection{Design of experiments}
To validate the theoretical results presented in Section \ref{Theory}, we design a set of experiments that compare the performance of the PERL framework with a standard NN model on the same vehicle trajectory prediction task. Specifically, we aim to empirically evaluate the performance of PERL about the required parameter size, convergence rate and the amount of training sample data needed to reach a desired performance level. These three aspects correspond directly to the theoretical advantages of PERL established in Section \ref{Theory}.

The dataset consists of car–following trajectories recorded in 2019 by the OpenACC project at Sweden’s AstaZero test track, with AVs serving as the following car. It was preprocessed by \citet{zhou2024ultra} through longitudinal trajectory extraction and cleaned to remove outliers and fill missing values. For model training, we use a subset of features: the speed and acceleration of both the leading and following vehicles, along with the inter-vehicle spacing. Given trajectory data from the past k=30 time steps (3 seconds, with a time interval of 0.1 seconds), the models are trained to predict the acceleration of the following vehicle at the next time step. 
In the PERL model, the Intelligent Driver Model (IDM) is chosen as the car-following physics model. The structure and parameter calibration of the IDM model are described in detail later in this section.  The IDM describes acceleration as a continuous function of speed, inter-vehicle gap, and speed difference, expressed as:
\begin{equation}
    a = a_{\max} \left[ 1 - \left(\dfrac{v}{v_0} \right)^\delta - \left(\dfrac{s^*(v, \Delta v)}{s}\right)^2 \right],
\end{equation}
where  
 \( a \) is the acceleration of the following vehicle,  
 \( v \) is the velocity of the following vehicle,  
 \( v_0 \) is the desired velocity,  
 \( a_{\max} \) is the maximum acceleration,  
 \( \delta \) is an acceleration exponent,  
 \( s \) is the actual inter-vehicle gap,  
 \( \Delta v = v_{\text{lead}} - v \) is the relative speed between the leading and following vehicles,  
 \( s(v, \Delta v) \) is the desired gap, given by:

\begin{equation}
    s(v, \Delta v) = s_0 + vT + \dfrac{v \Delta v}{2 \sqrt{a_{\max} b}},
\end{equation}
Where,  
 \( s_0 \) is the minimum gap at standstill,  
 \( T \) is the desired time headway,  
 \( b \) is the comfortable deceleration.  

The parameters of the IDM model are calibrated using the Monte Carlo method, by minimizing the MSE between the model outputs and empirical trajectory data. The calibrated model achieves an MSE of 0.0477, suggesting a close alignment with the observed car-following behavior. The calibrated parameters are as follows:

\begin{table}[h]
    \label{tab:IDM_params}
    \caption{Calibrated IDM parameters.}
    \centering
    \begin{tabular}{ccc}
        \toprule
        Parameter & Description & Calibrated Value \\
        \midrule
        \( v_0 \) & Desired velocity (m/s) & 23.058 \\
        \( a_{\max} \) & Maximum acceleration (m/s\(^2\)) & 0.572 \\
        \( b \) & Comfortable deceleration (m/s\(^2\)) & 2.601 \\
        \( s_0 \) & Minimum gap (m) & 1.605 \\
        \( T \) & Desired time headway (s) & 1.165 \\
        \bottomrule
    \end{tabular}
\end{table}
The residual learning component of PERL is implemented using a Long Short-Term Memory (LSTM) network, a widely-used model for time-series prediction. 
To ensure a fair comparison, the baseline NN model is also chosen as a standalone LSTM with the same architecture: two hidden layers with 32 neurons per layer and ReLU activation. 
Both PERL and the baseline LSTM model share the same input features and network architecture, but differ in their learning objectives. The LSTM model directly predicts the acceleration of the following vehicle, while PERL learns only the residual between the true acceleration and the output of the physics model. The experiments are structured to demonstrate the key advantages of PERL:
\begin{itemize}
  \item Reduction in the number of required parameters
  \item Improved convergence rate
  \item Reduced sample size for achieving the same prediction accuracy
\end{itemize}

\subsection{Results}

\subsubsection{Reduction in the number of required parameters}
\begin{figure}[ht]
    \centering
    \includegraphics[width=0.7\textwidth]{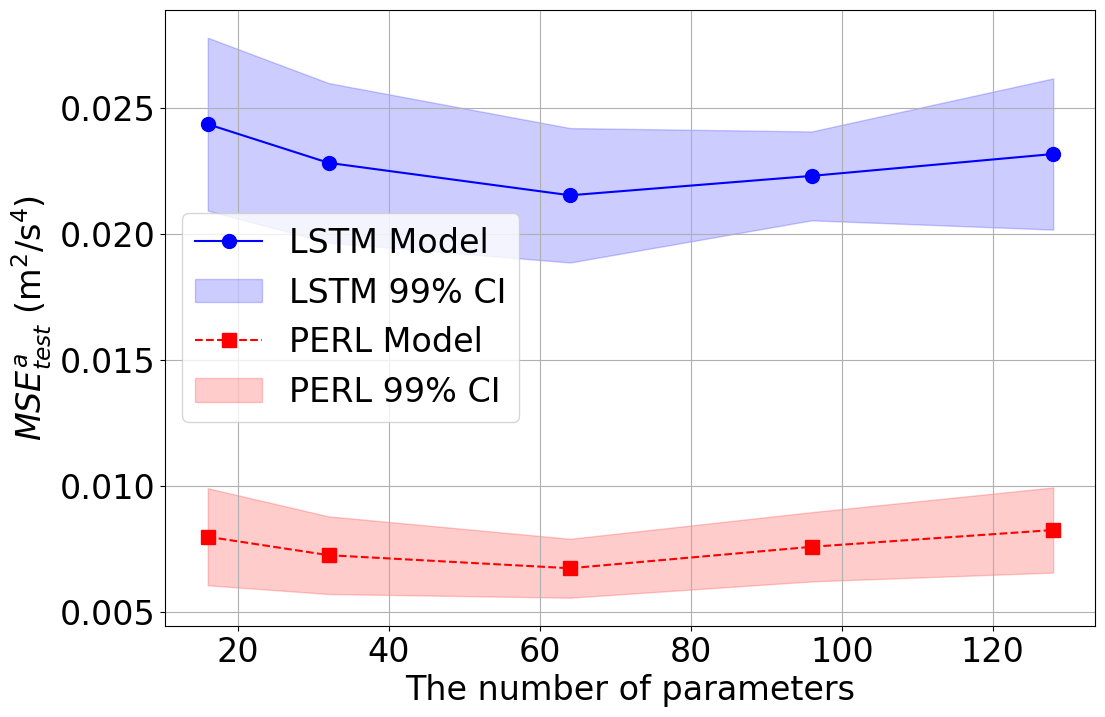}
    \caption{Comparison of LSTM and PERL Models Across Different Parameter Sizes.}
    \label{fig:parameters_experiment}
\end{figure}

To evaluate parameter efficiency, we vary the size of the hidden layers from 16 to 64 units and compare the test performance of the LSTM and PERL models, using 200 training samples and 200 training epochs. The prediction results of PERL and LSTM with different number of parameters are shown in Figure \ref{fig:parameters_experiment}. Across different parameter settings, the prediction accuracy of PERL consistently outperforms LSTM.
The performance improvement is especially notable when the number of parameters is small, confirming the theoretical result that PERL requires fewer parameters to reach a given accuracy due to the reduced complexity of the residual function. At larger sizes, overfitting begins to appear, particularly for LSTM.

\subsubsection{Improved convergence rate}
\begin{figure}[ht]
    \centering
    \includegraphics[width=0.7\textwidth]{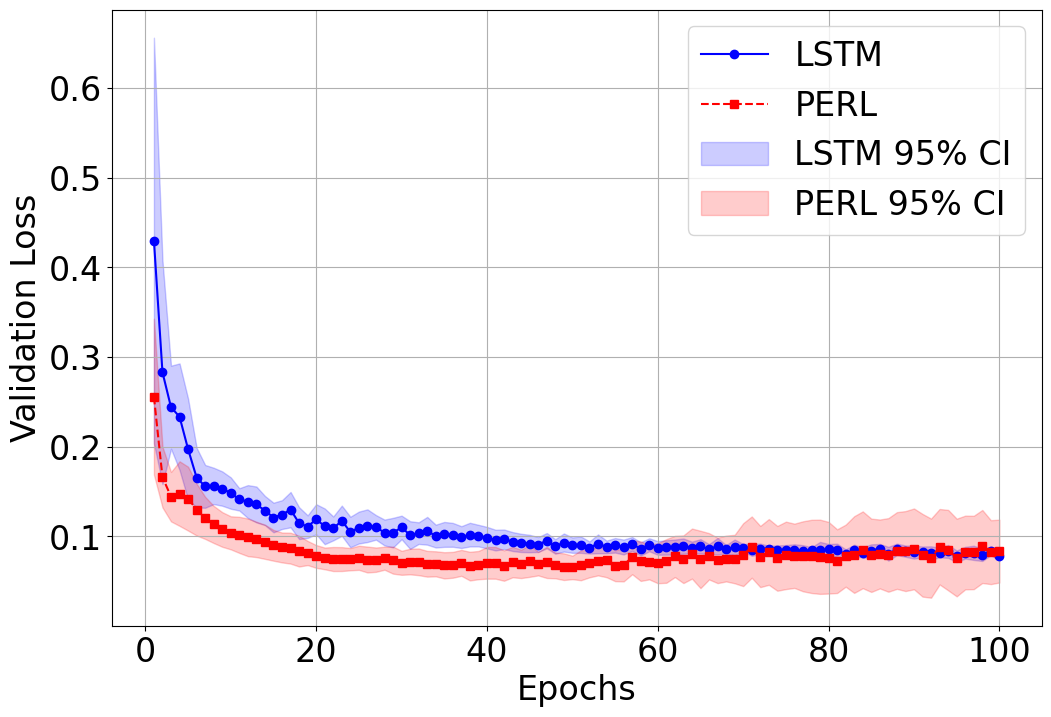}
    \caption{Validation loss comparison between LSTM and PERL models.}
    \label{fig:convergence_experiment}
\end{figure}

We next compare the convergence behavior of the two models. Figure~\ref{fig:convergence_experiment} shows the validation loss over 100 training epochs, averaged over 20 runs. PERL exhibits a faster drop in validation loss and reaches a lower final error than LSTM. This is consistent with our theoretical analysis: by learning a smoother residual function with a lower Lipschitz constant, PERL allows more efficient gradient-based optimization and faster convergence.

\subsubsection{Reduced sample size}
\begin{figure}[ht]
    \centering
    \includegraphics[width=0.75\textwidth]{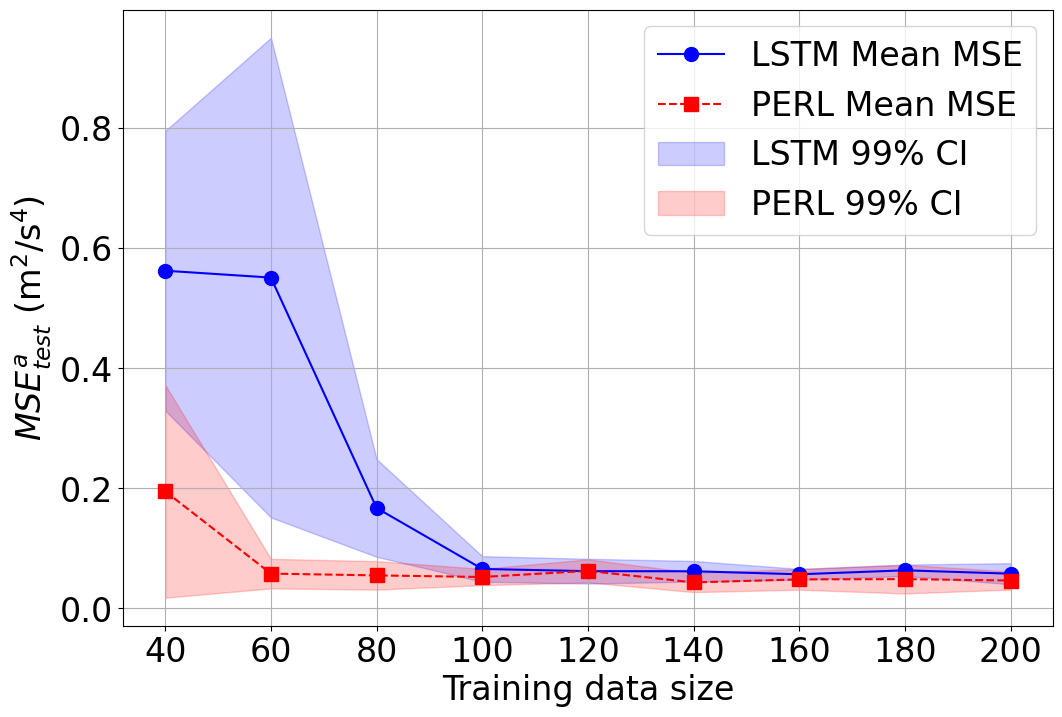}
    \caption{Comparison of LSTM and PERL Models Across Different Data Sizes}
    \label{fig:datasize_experiment}
\end{figure}

Finally, we evaluate the sample efficiency of the two models by varying the training set size from 20 to 200. As shown in Figure~\ref{fig:datasize_experiment}, PERL consistently outperforms LSTM.
 The difference is particularly evident when the training data size is small (e.g., fewer than 100 samples), where LSTM exhibits significantly higher variance and higher error, as shown by the larger 99\% CIs. In contrast, PERL maintains lower error and tighter confidence intervals, demonstrating its ability to generalize more effectively with limited data. This supports our theoretical result that PERL has a tighter generalization error bound and thus requires fewer samples to achieve comparable accuracy. As the training data size increases, the performance gap between the two models narrows, with both models converging to similar MSE values when provided with sufficient data (e.g., beyond 150 samples). However, this experiment still confirms that PERL enhances learning efficiency by reducing the sample size, making it valuable in scenarios where data availability is limited.

\section{Conclusion}
\label{conclusion}
This paper focused on Physics-Enhanced Residual Learning (PERL), a recently proposed method that integrates physics models into residual learning frameworks. While PERL has shown practical effectiveness in various applications, systematic theoretical analyses of its advantages have been lacking, leaving a significant research gap. To bridge this gap, we provided a theoretical foundation for PERL from three complementary perspectives: parameter size of NN, convergence rate, and statistical error bound. Specifically, our theoretical analyses demonstrated three key benefits of the PERL approach comparing with NN models:


\begin{itemize}
    \item \textbf{Reduction in the number of parameters} Compared with NN models,PERL simplifies the complexity of the target function. Consequently, the residual learning model in PERL requires fewer parameters to achieve comparable predictive accuracy, reduce the complexicity of computing. 
    
    \item \textbf{Improved convergence rate} Theoretical analysis showed that the residual function in PERL exhibits a lower Lipschitz constant, resulting in a faster convergence rates in the gradient descent optimization process, enhancing overall training efficiency.
    
    \item \textbf{Reduced sample complexity} By pre-processing the learning problem through a physics model, PERL requires fewer training samples to achieve the same level of estimation and generalization error as NN models.
\end{itemize}

We further validated our theoretical findings through three numerical experiments on vehicle trajectory prediction using real-world datasets. The results consistently aligned with the theoretical analyses, demonstrating that PERL outperforms LSTM model in terms of parameter efficiency, convergence speed, and learning efficiency. These findings reinforce the idea that integrating physics residual learning into NN training can improve both accuracy and interpretability while reducing computational costs.

The theoretical foundation for the PERL framework in this work can be further developed in several practical and methodological directions.
First, the theoretical insights developed in this paper can inform a broader class of PERL application beyond trajectory prediction. These include not only more complex prediction problems, such as PDE-constrained modeling or multi-agent interaction forecasting, but also control systems, where the residual learning component can be used to refine nominal physics-based controllers. In both contexts, new theoretical challenges arise, including the need to analyze stability, robustness, and performance under feedback. Future work may build on our current static results to develop dynamic generalizations of PERL using tools such as Lyapunov-based reasoning or robust control theory.
Second, from a systems perspective, PERL's efficiency, accuracy, and interpretability make it a compelling candidate for deployment in real-time transportation infrastructure. Applications include real-time control in connected corridors, edge-based safety analytics in digital twins, or embedded prediction models in autonomous vehicles. Future work may explore deploying PERL models under hardware constraints, studying trade-offs between prediction fidelity and latency, and integrating uncertainty quantification for risk-aware decision-making.
Third, the modular structure of PERL opens opportunities for hybrid modeling, where the physics and residual components can be decoupled and updated independently. This is particularly relevant for continual learning and online adaptation, as seen in evolving traffic environments. For example, when the physics model remains valid but the data distribution shifts (e.g., due to weather or regional changes), only the residual model may require updating. Investigating how to safely adapt PERL under such nonstationary conditions without retraining the full model is a promising direction for both theory and deployment.
\bibliographystyle{plainnat}  
\bibliography{references}

\begin{thebibliography}{52}
\providecommand{\natexlab}[1]{#1}
\providecommand{\url}[1]{\texttt{#1}}
\expandafter\ifx\csname urlstyle\endcsname\relax
  \providecommand{\doi}[1]{doi: #1}\else
  \providecommand{\doi}{doi: \begingroup \urlstyle{rm}\Url}\fi

\bibitem[Bartlett and Mendelson(2002)]{bartlett2002rademacher}
Peter~L Bartlett and Shahar Mendelson.
\newblock Rademacher and gaussian complexities: Risk bounds and structural results.
\newblock \emph{Journal of Machine Learning Research}, 3\penalty0 (Nov):\penalty0 463--482, 2002.

\bibitem[Bubeck et~al.(2015)]{bubeck2015convex}
S{\'e}bastien Bubeck et~al.
\newblock Convex optimization: Algorithms and complexity.
\newblock \emph{Foundations and Trends{\textregistered} in Machine Learning}, 8\penalty0 (3-4):\penalty0 231--357, 2015.

\bibitem[Chew et~al.(2024)Chew, Sender, Kaplan, Chandrasekaran, Chief~Elk, Browning, Kwak, Halls, and Afzal]{chew2024advancing}
Alex~K Chew, Matthew Sender, Zachary Kaplan, Anand Chandrasekaran, Jackson Chief~Elk, Andrea~R Browning, H~Shaun Kwak, Mathew~D Halls, and Mohammad Atif~Faiz Afzal.
\newblock Advancing material property prediction: using physics-informed machine learning models for viscosity.
\newblock \emph{Journal of Cheminformatics}, 16\penalty0 (1):\penalty0 31, 2024.

\bibitem[Cybenko(1989)]{cybenko1989approximation}
George Cybenko.
\newblock Approximation by superpositions of a sigmoidal function.
\newblock \emph{Mathematics of control, signals and systems}, 2\penalty0 (4):\penalty0 303--314, 1989.

\bibitem[Do et~al.(2019)Do, Vu, Vo, Liu, and Phung]{do2019effective}
Loan~NN Do, Hai~L Vu, Bao~Q Vo, Zhiyuan Liu, and Dinh Phung.
\newblock An effective spatial-temporal attention based neural network for traffic flow prediction.
\newblock \emph{Transportation research part C: emerging technologies}, 108:\penalty0 12--28, 2019.

\bibitem[Don{\`a} et~al.(2022)Don{\`a}, D{\'e}chelle, L{\'e}vy, and Gallinari]{dona2022constrainedICLR}
J{\'e}r{\'e}mie Don{\`a}, Marie D{\'e}chelle, Marina L{\'e}vy, and Patrick Gallinari.
\newblock Constrained physical-statistics models for dynamical system identification and prediction.
\newblock In \emph{ICLR 2022-The Tenth International Conference on Learning Representations}, 2022.

\bibitem[Doshi-Velez and Kim(2017)]{doshi2017towards}
Finale Doshi-Velez and Been Kim.
\newblock Towards a rigorous science of interpretable machine learning.
\newblock \emph{arXiv preprint arXiv:1702.08608}, 2017.

\bibitem[Du et~al.(2019)Du, Lee, Li, Wang, and Zhai]{du2019gradient}
Simon Du, Jason Lee, Haochuan Li, Liwei Wang, and Xiyu Zhai.
\newblock Gradient descent finds global minima of deep neural networks.
\newblock In \emph{International conference on machine learning}, pages 1675--1685. PMLR, 2019.

\bibitem[Emmanuel et~al.(2021)Emmanuel, Maupong, Mpoeleng, Semong, Mphago, and Tabona]{emmanuel2021survey}
Tlamelo Emmanuel, Thabiso Maupong, Dimane Mpoeleng, Thabo Semong, Banyatsang Mphago, and Oteng Tabona.
\newblock A survey on missing data in machine learning.
\newblock \emph{Journal of Big data}, 8:\penalty0 1--37, 2021.

\bibitem[Faroughi et~al.(2024)Faroughi, Pawar, Fernandes, Raissi, Das, Kalantari, and Kourosh~Mahjour]{faroughi2024physics}
Salah~A Faroughi, Nikhil~M Pawar, Celio Fernandes, Maziar Raissi, Subasish Das, Nima~K Kalantari, and Seyed Kourosh~Mahjour.
\newblock Physics-guided, physics-informed, and physics-encoded neural networks and operators in scientific computing: Fluid and solid mechanics.
\newblock \emph{Journal of Computing and Information Science in Engineering}, 24\penalty0 (4):\penalty0 040802, 2024.

\bibitem[Gurney(2018)]{gurney2018introduction}
Kevin Gurney.
\newblock \emph{An introduction to neural networks}.
\newblock CRC press, 2018.

\bibitem[Han et~al.(2015)Han, Pool, Tran, and Dally]{han2015learning}
Song Han, Jeff Pool, John Tran, and William Dally.
\newblock Learning both weights and connections for efficient neural network.
\newblock \emph{Advances in neural information processing systems}, 28, 2015.

\bibitem[He et~al.(2016)He, Zhang, Ren, and Sun]{he2016deep}
Kaiming He, Xiangyu Zhang, Shaoqing Ren, and Jian Sun.
\newblock Deep residual learning for image recognition.
\newblock In \emph{Proceedings of the IEEE conference on computer vision and pattern recognition}, pages 770--778, 2016.

\bibitem[He et~al.(2025)He, Gao, Xiao, Zhang, Wang, Wang, Luo, He, Sobanbab, Pan, et~al.]{he2025asap}
Tairan He, Jiawei Gao, Wenli Xiao, Yuanhang Zhang, Zi~Wang, Jiashun Wang, Zhengyi Luo, Guanqi He, Nikhil Sobanbab, Chaoyi Pan, et~al.
\newblock Asap: Aligning simulation and real-world physics for learning agile humanoid whole-body skills.
\newblock \emph{arXiv preprint arXiv:2502.01143}, 2025.

\bibitem[Hochreiter and Schmidhuber(1997)]{hochreiter1997long}
Sepp Hochreiter and J{\"u}rgen Schmidhuber.
\newblock Long short-term memory.
\newblock \emph{Neural computation}, 9\penalty0 (8):\penalty0 1735--1780, 1997.

\bibitem[Hoeffding(1963)]{hoeffding1963probability}
Wassily Hoeffding.
\newblock Probability inequalities for sums of bounded random variables.
\newblock \emph{Journal of the American statistical association}, 58\penalty0 (301):\penalty0 13--30, 1963.

\bibitem[Hoeffding(1994)]{hoeffding1994probability}
Wassily Hoeffding.
\newblock Probability inequalities for sums of bounded random variables.
\newblock \emph{The collected works of Wassily Hoeffding}, pages 409--426, 1994.

\bibitem[Hornik(1991)]{hornik1991approximation}
Kurt Hornik.
\newblock Approximation capabilities of multilayer feedforward networks.
\newblock \emph{Neural networks}, 4\penalty0 (2):\penalty0 251--257, 1991.

\bibitem[Hu et~al.(2024)Hu, Shukla, Karniadakis, and Kawaguchi]{hu2024tackling}
Zheyuan Hu, Khemraj Shukla, George~Em Karniadakis, and Kenji Kawaguchi.
\newblock Tackling the curse of dimensionality with physics-informed neural networks.
\newblock \emph{Neural Networks}, 176:\penalty0 106369, 2024.

\bibitem[Kang et~al.(2017)Kang, Lv, and Chen]{kang2017short}
Danqing Kang, Yisheng Lv, and Yuan-yuan Chen.
\newblock Short-term traffic flow prediction with lstm recurrent neural network.
\newblock In \emph{2017 IEEE 20th international conference on intelligent transportation systems (ITSC)}, pages 1--6. IEEE, 2017.

\bibitem[Ledoux and Talagrand(2013)]{ledoux2013probability}
Michel Ledoux and Michel Talagrand.
\newblock \emph{Probability in Banach Spaces: isoperimetry and processes}.
\newblock Springer Science \& Business Media, 2013.

\bibitem[Lipton(2018)]{lipton2018mythos}
Zachary~C Lipton.
\newblock The mythos of model interpretability: In machine learning, the concept of interpretability is both important and slippery.
\newblock \emph{Queue}, 16\penalty0 (3):\penalty0 31--57, 2018.

\bibitem[Liu and Ma(2024)]{liu2024navigating}
Qi~Liu and Wanjing Ma.
\newblock Navigating data corruption in machine learning: Balancing quality, quantity, and imputation strategies.
\newblock \emph{arXiv preprint arXiv:2412.18296}, 2024.

\bibitem[Long et~al.(2024)Long, Shi, and Li]{long2024PINN}
Keke Long, Xiaowei Shi, and Xiaopeng Li.
\newblock Physics-informed neural network for cross-dynamics vehicle trajectory stitching.
\newblock \emph{Transportation Research Part E: Logistics and Transportation Review}, 192:\penalty0 103799, 2024.

\bibitem[Long et~al.(2025)Long, Sheng, Shi, Li, Chen, and Ahn]{long2023physics}
Keke Long, Zihao Sheng, Haotian Shi, Xiaopeng Li, Sikai Chen, and Soyoung Ahn.
\newblock Physical enhanced residual learning (perl) framework for vehicle trajectory prediction.
\newblock \emph{Communications in Transportation Research}, 5:\penalty0 100166, 2025.

\bibitem[Mishra and Molinaro(2022)]{mishra2022estimates}
Siddhartha Mishra and Roberto Molinaro.
\newblock Estimates on the generalization error of physics-informed neural networks for approximating a class of inverse problems for pdes.
\newblock \emph{IMA Journal of Numerical Analysis}, 42\penalty0 (2):\penalty0 981--1022, 2022.

\bibitem[Mo et~al.(2021)Mo, Shi, and Di]{mo2021physics}
Zhaobin Mo, Rongye Shi, and Xuan Di.
\newblock A physics-informed deep learning paradigm for car-following models.
\newblock \emph{Transportation research part C: emerging technologies}, 130:\penalty0 103240, 2021.

\bibitem[Moody(1991)]{moody1991effective}
John Moody.
\newblock The effective number of parameters: An analysis of generalization and regularization in nonlinear learning systems.
\newblock \emph{Advances in neural information processing systems}, 4, 1991.

\bibitem[Nesterov(2013)]{nesterov2013introductory}
Yurii Nesterov.
\newblock \emph{Introductory lectures on convex optimization: A basic course}, volume~87.
\newblock Springer Science \& Business Media, 2013.

\bibitem[Pan et~al.(2024)Pan, Xiao, and Shen]{pan2024ro}
Renbin Pan, Feng Xiao, and Minyu Shen.
\newblock ro-pinn: A reduced order physics-informed neural network for solving the macroscopic model of pedestrian flows.
\newblock \emph{Transportation Research Part C: Emerging Technologies}, 163:\penalty0 104658, 2024.

\bibitem[Raissi et~al.(2017)Raissi, Perdikaris, and Karniadakis]{raissi2017physics}
Maziar Raissi, Paris Perdikaris, and George~Em Karniadakis.
\newblock Physics informed deep learning (part i): Data-driven solutions of nonlinear partial differential equations.
\newblock \emph{arXiv preprint arXiv:1711.10561}, 2017.

\bibitem[Raissi et~al.(2019)Raissi, Perdikaris, and Karniadakis]{raissi2019physics}
Maziar Raissi, Paris Perdikaris, and George~E Karniadakis.
\newblock Physics-informed neural networks: A deep learning framework for solving forward and inverse problems involving nonlinear partial differential equations.
\newblock \emph{Journal of Computational physics}, 378:\penalty0 686--707, 2019.

\bibitem[Ruder(2016)]{ruder2016overview}
Sebastian Ruder.
\newblock An overview of gradient descent optimization algorithms.
\newblock \emph{arXiv preprint arXiv:1609.04747}, 2016.

\bibitem[Rumelhart et~al.(1986)Rumelhart, Hinton, and Williams]{rumelhart1986learning}
David~E Rumelhart, Geoffrey~E Hinton, and Ronald~J Williams.
\newblock Learning representations by back-propagating errors.
\newblock \emph{nature}, 323\penalty0 (6088):\penalty0 533--536, 1986.

\bibitem[Said and Erradi(2019)]{said2019deep}
Ahmed~Ben Said and Abdelkarim Erradi.
\newblock Deep-gap: A deep learning framework for forecasting crowdsourcing supply-demand gap based on imaging time series and residual learning.
\newblock In \emph{2019 IEEE International Conference on Cloud Computing Technology and Science (CloudCom)}, pages 279--286. IEEE, 2019.

\bibitem[Shalev-Shwartz and Ben-David(2014)]{shalev2014understanding}
Shai Shalev-Shwartz and Shai Ben-David.
\newblock \emph{Understanding machine learning: From theory to algorithms}.
\newblock Cambridge university press, 2014.

\bibitem[Shi et~al.(2023)Shi, Zhou, Wu, Chen, Ran, and Nie]{shi2023physics}
Haotian Shi, Yang Zhou, Keshu Wu, Sikai Chen, Bin Ran, and Qinghui Nie.
\newblock Physics-informed deep reinforcement learning-based integrated two-dimensional car-following control strategy for connected automated vehicles.
\newblock \emph{Knowledge-Based Systems}, 269:\penalty0 110485, 2023.

\bibitem[Shi et~al.(2022)Shi, Wu, Shi, Zhou, and Ran]{shi2022integrated}
Kunsong Shi, Yuankai Wu, Haotian Shi, Yang Zhou, and Bin Ran.
\newblock An integrated car-following and lane changing vehicle trajectory prediction algorithm based on a deep neural network.
\newblock \emph{Physica A: Statistical Mechanics and its Applications}, 599:\penalty0 127303, 2022.

\bibitem[Shultzman et~al.(2023)Shultzman, Azar, Rodrigues, and Eldar]{shultzman2023generalization}
Avner Shultzman, Eyar Azar, Miguel~RD Rodrigues, and Yonina~C Eldar.
\newblock Generalization and estimation error bounds for model-based neural networks.
\newblock \emph{arXiv preprint arXiv:2304.09802}, 2023.

\bibitem[Trottenberg et~al.(2001)Trottenberg, Oosterlee, and Schuller]{trottenberg2001multigrid}
Ulrich Trottenberg, Cornelius~W Oosterlee, and Anton Schuller.
\newblock \emph{Multigrid methods}.
\newblock Academic press, 2001.

\bibitem[Vaswani et~al.(2017)Vaswani, Shazeer, Parmar, Uszkoreit, Jones, Gomez, Kaiser, and Polosukhin]{vaswani2017attention}
Ashish Vaswani, Noam Shazeer, Niki Parmar, Jakob Uszkoreit, Llion Jones, Aidan~N Gomez, {\L}ukasz Kaiser, and Illia Polosukhin.
\newblock Attention is all you need.
\newblock \emph{Advances in neural information processing systems}, 30, 2017.

\bibitem[Wang et~al.(2020{\natexlab{a}})Wang, Mo, and Zhao]{wang2020deep}
Shenhao Wang, Baichuan Mo, and Jinhua Zhao.
\newblock Deep neural networks for choice analysis: Architecture design with alternative-specific utility functions.
\newblock \emph{Transportation Research Part C: Emerging Technologies}, 112:\penalty0 234--251, 2020{\natexlab{a}}.

\bibitem[Wang et~al.(2020{\natexlab{b}})Wang, Wang, and Zhao]{wang2020deep1}
Shenhao Wang, Qingyi Wang, and Jinhua Zhao.
\newblock Deep neural networks for choice analysis: Extracting complete economic information for interpretation.
\newblock \emph{Transportation Research Part C: Emerging Technologies}, 118:\penalty0 102701, 2020{\natexlab{b}}.

\bibitem[Wang and Zhong(2024)]{wang2024pinn}
Yifan Wang and Linlin Zhong.
\newblock Nas-pinn: neural architecture search-guided physics-informed neural network for solving pdes.
\newblock \emph{Journal of Computational Physics}, 496:\penalty0 112603, 2024.

\bibitem[Welch et~al.(1995)Welch, Bishop, et~al.]{welch1995introduction}
Greg Welch, Gary Bishop, et~al.
\newblock \emph{An introduction to the Kalman filter}.
\newblock Chapel Hill, NC, USA, 1995.

\bibitem[Yarotsky(2017)]{yarotsky2017error}
Dmitry Yarotsky.
\newblock Error bounds for approximations with deep relu networks.
\newblock \emph{Neural networks}, 94:\penalty0 103--114, 2017.

\bibitem[Yuan et~al.(2021)Yuan, Zhang, Yang, and Zhe]{yuan2021macroscopic}
Yun Yuan, Zhao Zhang, Xianfeng~Terry Yang, and Shandian Zhe.
\newblock Macroscopic traffic flow modeling with physics regularized gaussian process: A new insight into machine learning applications in transportation.
\newblock \emph{Transportation Research Part B: Methodological}, 146:\penalty0 88--110, 2021.

\bibitem[Zhang et~al.(2017)Zhang, Zheng, and Qi]{zhang2017deep}
Junbo Zhang, Yu~Zheng, and Dekang Qi.
\newblock Deep spatio-temporal residual networks for citywide crowd flows prediction.
\newblock In \emph{Proceedings of the AAAI conference on artificial intelligence}, volume~31, 2017.

\bibitem[Zhang et~al.(2024)Zhang, Huang, Zhou, Shi, Long, and Li]{zhang2024online}
Peng Zhang, Heye Huang, Hang Zhou, Haotian Shi, Keke Long, and Xiaopeng Li.
\newblock Online adaptive platoon control for connected and automated vehicles via physics enhanced residual learning.
\newblock \emph{arXiv preprint arXiv:2412.20680}, 2024.

\bibitem[Zhou et~al.(2024)Zhou, Ma, Liang, Li, and Qu]{zhou2024ultra}
Hang Zhou, Ke~Ma, Shixiao Liang, Xiaopeng Li, and Xiaobo Qu.
\newblock Ultra-av: A unified longitudinal trajectory dataset for automated vehicle.
\newblock \emph{arXiv preprint arXiv:2406.00009}, 2024.

\bibitem[Zhou et~al.(2017)Zhou, Qu, and Li]{zhou2017recurrent}
Mofan Zhou, Xiaobo Qu, and Xiaopeng Li.
\newblock A recurrent neural network based microscopic car following model to predict traffic oscillation.
\newblock \emph{Transportation research part C: emerging technologies}, 84:\penalty0 245--264, 2017.

\bibitem[Zinkevich(2003)]{zinkevich2003online}
Martin Zinkevich.
\newblock Online convex programming and generalized infinitesimal gradient ascent.
\newblock In \emph{Proceedings of the 20th international conference on machine learning (icml-03)}, pages 928--936, 2003.

\end{thebibliography}

\appendix
\clearpage
\section{Proof of Theorem \ref{theorem: number of parameters}}
\label{pf: number of parameters}
\noindent Consider a family of functions \( \mathcal{F} \) defined on the interval \([a, b]\), where each \( f \in \mathcal{F} \) is Lipschitz continuous with the same Lipschitz constant \( L \). We aim to show that for any \( f \in \mathcal{F} \), it is possible to approximate \( f \) using at most \( N \) linear segments, such that the total approximation error does not exceed \( \varepsilon \), where
\[
P = \left\lceil \dfrac{L(b - a)^2}{4 \varepsilon} \right\rceil.
\]

To construct the approximation, we divide the interval \([a, b]\) into \( P \) equal subintervals of length
\[
\Delta x = \dfrac{b - a}{P}.
\]
On each subinterval \([x_i, x_{i+1}]\), approximate \( f(x) \) by the linear function \( \hat{f}(x) \) that interpolates \( f \) at the endpoints \( x_i \) and \( x_{i+1} \).

Since each \( f \in \mathcal{F} \) is Lipschitz continuous with constant \( L \), the maximum error on each subinterval is bounded by
\[
E_{\text{max}} = \dfrac{L \Delta x}{2}.
\]
For any \( f \in \mathcal{F} \), the error function \( e(x) = f(x) - \hat{f}(x) \) on each subinterval increases from 0 at \( x_i \) to \( E_{\text{max}} \) at the midpoint and then decreases back to 0 at \( x_{i+1} \). Therefore, \( e(x) \) forms a triangle over each subinterval.

The integral of the absolute error over a single subinterval is the area of this triangle:
\begin{align*}
E_{\text{interval}}
&= \frac{1}{2} \times \text{base} \times \text{height} \\
&= \frac{1}{2} \times \Delta x \times E_{\text{max}} \\
&= \frac{1}{2} \times \Delta x \times \left( \frac{L \Delta x}{2} \right) \\
&= \frac{L (\Delta x)^2}{4}.
\end{align*}
The total error over all \( P \) subintervals is then
\[
E_{\text{total}} = P \times E_{\text{interval}} = P \times \dfrac{L (\Delta x)^2}{4}.
\]
Substituting \( \Delta x = \dfrac{b - a}{P} \) gives
\[
E_{\text{total}} = P\times \dfrac{L \left( \dfrac{b - a}{P} \right)^2}{4} = \dfrac{L (b - a)^2}{4P}.
\]
To ensure that the total error does not exceed \( \varepsilon \), we require
\[
E_{\text{total}} \leq \varepsilon \quad \Rightarrow \quad \dfrac{L (b - a)^2}{4P} \leq \varepsilon.
\]
Solving for \( P \), we get
\[P \geq \dfrac{L (b - a)^2}{4 \varepsilon}.
\]
Thus, the smallest integer satisfying this inequality is
\[
P = \left\lceil \dfrac{L (b - a)^2}{4 \varepsilon} \right\rceil.
\]
This shows that at most \( P \) segments are sufficient for any \( f \in \mathcal{F} \) to achieve the desired error tolerance.

Finally, consider the case where \( f(x) = Lx + c \), a linear function with slope \( L \). For this function, the total error exactly reaches the threshold \( \varepsilon \) when we use \( P \) segments, as given by the formula. Therefore, \( P \) is not only an upper bound but also the smallest possible value that meets the approximation error requirement, confirming that \(P \) is the supremum.

This completes the proof that \( P \) is the supremum on the number of segments required for the approximation of any \( f \in \mathcal{F} \) to achieve the specified error tolerance.


\section{Proof of Lemma 
\label{pf:integral}
\ref{lemma: integral}}
\noindent We estimate the sum using an integral bound:
\begin{equation}
\sum_{t=1}^T \dfrac{1}{\sqrt{t}} < 1 + \int_1^T \dfrac{dt}{\sqrt{t}}.
\end{equation}
Evaluating the integral, we obtain:
\begin{equation}
1 + 2(\sqrt{T} - 1) = 2\sqrt{T} - 1 < 2\sqrt{T}.
\end{equation}

\section{Proof of Corollary \ref{Corollary: non-constant step size convergence rate bound}}
\label{pf:non-constant-step-size-convergence-rate-bound}
We denote by \(x^*=\arg\min_x f(x)\) a global minimizer, and write the GD update:
\[
x^{t+1} \;=\; x^t \;-\;\eta_t\,\nabla f\bigl(x^t\bigr).
\]
Expanding the squared distance to \(x^*\) gives
\begin{align*}
\|x^{t+1}-x^*\|^2
&= \|x^t-x^*\|^2
- 2\,\eta_t\,\nabla f(x^t)^\top(x^t-x^*) \\
&\quad+ \eta_t^2\|\nabla f(x^t)\|^2
\end{align*}

By convexity, \(\nabla f(x^t)^\top(x^t-x^*)\ge f(x^t)-f(x^*)\), and since \(\|\nabla f(x^t)\|\le L\), we have
\[
f(x^t)-f(x^*)
\;\le\;
\frac{\|x^t-x^*\|^2 - \|x^{t+1}-x^*\|^2}{2\,\eta_t}
\;+\;\frac{\eta_t\,L^2}{2}.
\]
Summing over \(t=1,\dots,T\) yields
\begin{align*}
\sum_{t=1}^T\bigl(f(x^t)-f(x^*)\bigr)
&\le \sum_{t=1}^T \frac{\|x^t-x^*\|^2 - \|x^{t+1}-x^*\|^2}{2\,\eta_t} \\
&\quad+ \frac{L^2}{2}\sum_{t=1}^T \eta_t
\end{align*}

We now bound each term separately:

1.  
\begin{align*}
\sum_{t=1}^T \frac{\|x^t-x^*\|^2 - \|x^{t+1}-x^*\|^2}{2\,\eta_t} 
\\
= \frac{\|x^1-x^*\|^2}{2\,\eta_1} 
- \frac{\|x^{T+1}-x^*\|^2}{2\,\eta_T} 
\\
\quad + \sum_{t=2}^T \frac{\|x^t-x^*\|^2}{2}
\left(\frac{1}{\eta_t} - \frac{1}{\eta_{t-1}}\right)
\end{align*}

Since \(\|x^t-x^*\|\le B\) and \(\eta_t^{-1}=\sqrt{t}\), this is
\begin{align*}
&\le \frac{B^2}{2}\bigl(\sqrt{1}\bigr)
  + \frac{B^2}{2}\sum_{t=2}^T\bigl(\sqrt{t}-\sqrt{t-1}\bigr) \\
&= \frac{B^2}{2} + \frac{B^2}{2}\bigl(\sqrt{T}-1\bigr) \\
&= \frac{B^2 \sqrt{T}}{2}.
\end{align*}
2. 
   By Lemma~\ref{lemma: integral}, \(\sum_{t=1}^T\eta_t = \sum_{t=1}^T 1/\sqrt{t}\le 2\sqrt{T}\).  Hence
   \[
   \frac{L^2}{2}\sum_{t=1}^T \eta_t
   \;\le\;
   L^2 \sqrt{T}.
   \]

Combining these bounds gives
\begin{align*}
\sum_{t=1}^T\bigl(f(x^t)-f(x^*)\bigr)
&\le \frac{B^2}{2} + \frac{B^2}{2}\bigl(\sqrt{T}-1\bigr) + L^2 \sqrt{T} \\
&= \left(\frac{B^2}{2} + L^2\right)\sqrt{T}.
\end{align*}
Dividing by \(T\) and using \(\eta_t=1/\sqrt{t}\) concludes the proof:
\[
\frac1T\sum_{t=1}^T\bigl(f(x^t)-f(x^*)\bigr)
\;\le\;
\frac{\tfrac{B^2}{2}+L^2}{\sqrt T}.
\]

\section{Proof of Lemma \ref{Inequalities}}
\label{pf: Inequality}
\noindent We want to bound \( R(f) - \hat{R}(f) \) using the inequality:

$$\mathbb{P}(|\hat{R}(f) - R(f)| > t) \leq 2 e^{-2nt^2 / c_f^2}.$$. 

 \( f^\star = \arg\min_{f \in \mathcal{F}} R(f) \) and \( \hat{f} = \arg\min_{f \in \mathcal{F}} \hat{R}(f) \).

According to the inequality, we have with probability at least \( 1 - \delta \):
\[
R(f) \leq \hat{R}(f) + t
\]

for any function \( f \in \mathcal{F} \). Setting \( f = \hat{f} \), we obtain:
\[
R(\hat{f}) \leq \hat{R}(\hat{f}) + t.
\]

By the definition of \( \hat{f} \) as the minimizer of \( \hat{R}(f) \), we have:
\[
\hat{R}(\hat{f}) \leq \hat{R}(f^\star).
\]

Again, applying the inequality, we know with probability at least \( 1 - \delta \):
\[
\hat{R}(f^\star) \leq R(f^\star) + t.
\]

Putting it all together, with probability at least \( 1 - \delta \) we have:
\[
R(\hat{f}) \leq \hat{R}(\hat{f}) + t \leq \hat{R}(f^\star) + t \leq R(f^\star) + 2t.
\]

\section{Proof of Theorem \ref{theorem: estimation error}}
\label{pf: estimation error}
\noindent We start by rewriting the probability expression:
\begin{align*}
\mathbb{P}\big(R(\hat{f}) - R(f^*) > 2t\big) 
&= \mathbb{P}\Big(
      R(\hat{f}) - \hat{R}(\hat{f}) \\
&\hspace{2em}
      + \hat{R}(\hat{f}) - R(f^*) > 2t
    \Big) \\
&= \mathbb{P}\Big(
      \{R(\hat{f}) - \hat{R}(\hat{f}) > t\} \\
&\hspace{2em}
      \cup \{\hat{R}(\hat{f}) - R(f^*) > t\}
    \Big) \\
&\le \mathbb{P}\big(R(\hat{f}) - \hat{R}(\hat{f}) > t\big) \\
&\quad + \mathbb{P}\big(\hat{R}(\hat{f}) - R(f^*) > t\big) \\
&\le \delta + \delta \\
&= 2\delta \\
&= 4 e^{-2nt^2 / c_f^2}.
\end{align*}
Let \( 2t = \epsilon \), so that \( t = \dfrac{\epsilon}{2} \), and substitute \( \epsilon \) into the equation to complete the proof.

\section{Proof of Lemma \ref{lemma: Loss Lip constant}}
\begin{proof}
For any \(s, s' \in \Omega\),
\begin{align*}
|\ell(s; \theta) - \ell(s'; \theta)| 
&= \bigl|(f(s) - g(s))^2 - (f(s') - g(s'))^2\bigr| \\[4pt]
&= \bigl|(f(s) - f(s')) \\
&\quad\cdot \big(f(s) + f(s') - 2g(s)\big) \\[2pt]
&\quad+ (g(s) - g(s')) \\
&\quad\cdot \big(2f(s') - g(s) - g(s')\big) \bigr| \\[4pt]
&\le L\,\|s - s'\| \cdot 4C
    + L_g\,\|s - s'\| \cdot 4C \\[2pt]
&= 4C\,(L + L_g)\,\|s - s'\|.
\end{align*}

Since \(g(s)\) is the ground truth function and not subject to Lipschitz variation, we set \(L_g = 0\) and obtain
\[
|\ell(s; \theta) - \ell(s'; \theta)| 
\;\le\;4C\,L\,\|s - s'\|.
\]
Hence \(L_\ell = 4CL\).
\end{proof}

\section{Proof of lemma \ref{lemma:contraction_loss}}

Apply the lemma above with $\phi_i(f) = \varphi(z_i(f))$ and $\psi_i(f) = L z_i(f)$.
Since $\ell$ is an $L$-Lipschitz function such that for any two values $z$ and $z'$, the function $\ell$ satisfies:
\[
|\ell(z) - \ell(z')| \leq L |z - z'|.
\]

Using this substitution in the lemma, we obtain the inequality:
\[
\mathbb{E} \left[ \sup_{f \in \mathcal{F}} \sum_{i=1}^n \sigma_i \ell(z_i(f)) \right] \leq L \, \mathbb{E} \left[ \sup_{f \in \mathcal{F}} \sum_{i=1}^n \sigma_i z_i(f) \right].
\]

\section{Proof of Theorem \ref{thm:generalization_w/o loss}}
\label{pf: generalization error}
We start from the general Rademacher-based generalization bound stated in Theorem \ref{theorem:generalization}: for an $L_\ell$-Lipschitz loss function $\ell$ bounded in $[0, c]$, we have with probability at least $1 - \delta$:
\begin{equation}
R(f) - \hat{R}_n(f) \leq 2 \mathcal{R}_n(\ell \circ \mathcal{F}) + c \sqrt{\dfrac{\log(1/\delta)}{2n}}.
\label{eq:gen_rademacher_lipschitz}
\end{equation}

To apply this result, we need to upper bound the Rademacher complexity of the composed class $\ell \circ \mathcal{F} = \{s \mapsto \ell(s; \theta) \mid f \in \mathcal{F} \}$. 

Lemma~\ref{lemma: Loss Lip constant} shows that when the prediction function $f$ is $L$-Lipschitz and both $f(s)$ and $g(s)$ are bounded by $C$, then the loss function $\ell(s; \theta) = (f(s \mid \theta) - g(s))^2$ is $L_\ell = 4CL$-Lipschitz continuous with respect to $s$.

Next, Lemma~\ref{lemma:contraction} provides a general inequality that allows us to relate the Rademacher complexity of a Lipschitz-transformed function class to the original one. Applying this lemma with transformation $\phi_i(f) = \ell(z_i(f))$ and noting that $\ell$ is $L_\ell$-Lipschitz, we obtain from Lemma~\ref{lemma:contraction_loss}
\[
\mathcal{R}_n(\ell \circ \mathcal{F}) \leq L_\ell \cdot \mathcal{R}_n(\mathcal{F}) = 4CL \cdot \mathcal{R}_n(\mathcal{F}).
\]

Substituting this into inequality~\eqref{eq:gen_rademacher_lipschitz}, we get:
\begin{align*}
R(f) - \hat{R}_n(f)
&\le 2 \cdot 4CL \cdot \mathcal{R}_n(\mathcal{F})
  + c \sqrt{\frac{\log(1/\delta)}{2n}} \\
&= 8CL \cdot \mathcal{R}_n(\mathcal{F})
  + c \sqrt{\frac{\log(1/\delta)}{2n}}.
\end{align*}

which proves the claim.

\end{document}